\pdfoutput=1

\documentclass[11pt]{article}

\usepackage[preprint]{acl}

\usepackage{times}
\usepackage{latexsym}
\usepackage{amsmath,amssymb,amsfonts}
\usepackage{booktabs} 
\usepackage{multirow}
\usepackage[normalem]{ulem}
\usepackage{enumitem}
\usepackage{subcaption}
\usepackage{algorithm}
\usepackage{tabularx} 
\usepackage{algpseudocode}
\usepackage{array}
\usepackage{listings}
\usepackage{xcolor}
\usepackage{courier}
\usepackage[T1]{fontenc}
\usepackage[utf8]{inputenc}

\usepackage{microtype}

\usepackage{inconsolata}

\usepackage{graphicx}
\usepackage{url}
\usepackage[normalem]{ulem}
\usepackage{xcolor}
\usepackage{enumitem}
\makeatletter
\def\thanks#1{\protected@xdef\@thanks{\@thanks\protect\footnotetext{#1}}}
\makeatother

\title{
DioR: Adaptive Cognitive Detection and Contextual Retrieval Optimization for Dynamic Retrieval-Augmented Generation}

\author{Hanghui Guo$^{1,2}$, Jia Zhu$^{1*}$, {\bf Shimin Di$^{3*}$, Weijie Shi$^{4*}$, Zhangze Chen$^1$, Jiajie Xu$^5$} \thanks{First author.} \thanks{ghh1125@zjnu.edu.cn} \thanks{} \thanks{$^*$Corresponding author.} \thanks{jiazhu@zjnu.edu.cn} \thanks{shimin.di@seu.edu.cn} \thanks{wshiah@connect.ust.hk}\\
$^1$Zhejiang Key Laboratory of Intelligent Education Technology and Application, Zhejiang Normal University \\
$^2$School of Computer Science and Technology, Zhejiang Normal University \\ $^3$School of Computer Science and Engineering, Southeast University \\ $^4$Department of Computer Science and Engineering, Hong Kong University of Science and Technology \\ $^5$School of Computer Science and Technology, Soochow University}

\begin{document}
\maketitle
\begin{abstract}
Dynamic Retrieval-augmented Generation (RAG) has shown great success in mitigating hallucinations in large language models (LLMs) during generation. However, existing dynamic RAG methods face significant limitations in two key aspects:
\textbf{\textit{1) Lack of an effective mechanism to control retrieval triggers, and 2) Lack of effective scrutiny of retrieval content.}}
To address these limitations, we propose an innovative dynamic RAG method,  
DioR (Adaptive Cognitive \textbf{D}etect\textbf{i}on and C\textbf{o}ntextual \textbf{R}etrieval Optimization), which consists of two main components: 
\textit{adaptive cognitive detection} and \textit{contextual retrieval optimization},
specifically 
designed to determine when retrieval is needed and what to retrieve for LLMs is useful.
Experimental results demonstrate that DioR achieves superior performance on all tasks, demonstrating the effectiveness of our work. 
\end{abstract}

\section{Introduction}

Large language models (LLMs) have demonstrated remarkable capabilities across various generative tasks, such as text creation, dialogue generation, and content summarization \cite{raffel2020exploring, brown2020language, achiam2023gpt, chan2024rq}. However, LLMs remain inherently static and lack the ability to learn in real-time \cite{mallen2022not, xing2024understanding, kumar2024large}. As a result, they cannot incorporate up-to-date information, which may generate inaccurate or even fabricated content when encountering unfamiliar scenarios \cite{maynez2020faithfulness}. This phenomenon is commonly referred to as \textit{``hallucination''}.

\begin{figure}[t]
  \centering
\includegraphics[width=\linewidth]{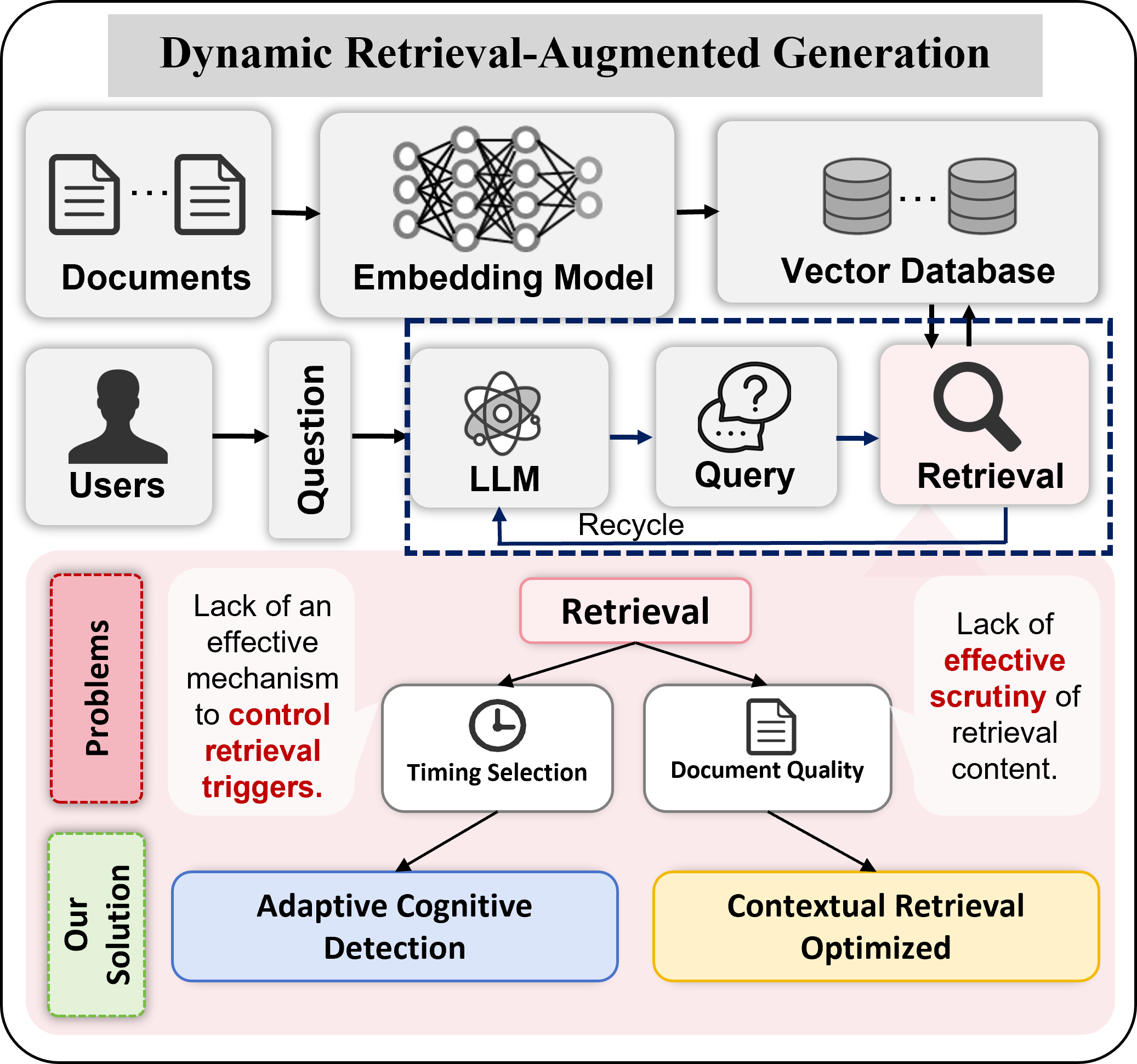} 
  \caption {Dynamic RAG and its limitations about ``Timing Selection'' and ``Document Quality'', as well as our solution to address the ``Retrieval'' limitation.}
  \label{fig:rag}
\end{figure}

Retrieval-Augmented Generation (RAG) has gained attention as an innovative approach to reducing hallucinations in LLM. By integrating external knowledge bases during the generation process and leveraging the contextual learning capabilities of LLMs, RAG effectively minimizes the generation of erroneous information, enhancing the reliability and precision of LLM generation \cite{jiang2022retrieval, jiang2024longrag, niu2023ragtruth, Zhu_Guo_Shi_Chen_DeMeo_2025}.

Conventional RAG methods are a single-turn retrieval process, where relevant documents are extracted based on the LLM's input query and used for content generation \cite{ye2024boosting, roy2024learning, zhang2024question}. While effective for simpler tasks, this method struggles with complex tasks or extensive text generation \cite{he2024retrieving}. 
These highlight the need for advanced retrieval mechanisms to handle these complex scenarios.

In contrast, Dynamic RAG effectively alleviates this issue by enhancing the quality of LLM outputs through multi-turn retrieval during 
the generation process \cite{tayal2024dynamic, DBLP:conf/nips/YangSXSBCCGJJKM24, DBLP:conf/acl/SuTA0024}.
Specifically, it consists of two main steps: 1) Selecting the appropriate retrieval timing, and 2) 
Constructing the appropriate query once retrieval is triggered.
This approach effectively addresses the shortcomings of conventional RAG in handling more complex problems.




However, as shown in Figure \ref{fig:rag}, existing dynamic RAG methods exhibit significant limitations from two perspectives, specifically:
\begin{enumerate}[leftmargin=*]
    \item 
    \textit{\textbf{Lack of an effective mechanism to control retrieval triggers:}} Current dynamic RAG strategies typically rely on static, predefined rules to trigger the retrieval module, such as when the generation probability of a token falls below a predefined threshold. However, a low generation probability only indicates low confidence in the token, which does not necessarily imply hallucination after LLM generation text. Moreover, existing RAG strategies often trigger retrieval only after hallucinations occur during the LLM text generation process. If the existence mechanism could predict whether the LLM is capable of answering a question in advance and trigger retrieval accordingly before generating, hallucinations might be avoided when generating.

    \item 
    \textit{\textbf{Lack of effective scrutiny of retrieval content:}} The current dynamic RAG method performs single-batch retrieval in each round and also relies on the confidence scores of the most recent sentences' tokens generated by the LLM to determine the final retrieval keyword, without fully considering the overall contextual requirements of the task. This can lead to the retrieval of documents that are not entirely relevant, as well as the introduction of noisy data. In addition, single-batch retrieval often focuses on limited aspects of the context, and the retrieved information has high redundancy, leading to repeated retrievals and increased computational costs. Moreover, retrieving documents with excessively long content can significantly hinder the LLM’s ability to understand, leading to information overload and impacting the model’s ability to extract key information from documents and perform effective reasoning.
    
\end{enumerate}





To bridge these gaps, we propose a novel dynamic RAG method, named \textbf{DioR} (Adaptive Cognitive \textbf{D}etect\textbf{i}on and C\textbf{o}ntextual \textbf{R}etrieval Optimization). 
DioR consists of two main components: 
\textit{adaptive cognitive detection}, and \textit{ contextual retrieval optimization}. Specifically:

In terms of \textit{adaptive cognitive detection}, we utilize the Wiki dataset to construct two hallucination detection datasets and train two classifiers: Early Detection and Real-time Detection. The Early Detection Classifier assesses the model's ability to answer questions independently, while Real-time Detection monitors for hallucinations during the generation process. Once Early Detection finds LLM has no confidence in its response or Real-time Detection identifies hallucinations, we present \textit{contextual retrieval optimization} to step-by-step retrieve documents. We optimize the priority of query keywords through contextual analysis for retrieval. Retrieved documents are then used to capture new concepts to refine query keywords for more relevant and precise retrieval in later steps. Additionally, we use a sentence-level chunking module to reduce the impact of long texts, enhancing model understanding and inference performance.


The main contributions are listed as follows:


\begin{itemize}[leftmargin=*]   

\item We propose DioR, a novel dynamic RAG method that addresses two key limitations of existing methods: 1) Lack of an effective mechanism to control retrieval triggers, and 2) Lack of effective scrutiny of retrieval content.

\item We constructed and trained both early detection and real-time detection classifiers to determine the optimal retrieval trigger timing based on the internal state of the LLM before and during the text generation process. Additionally, we adopted contextual retrieval optimization to enhance both the retrieval process and the quality of the documents, thereby improving the LLM's reasoning generation capabilities.

\item  Our experiments demonstrate that DioR surpasses popular methods in four knowledge-intensive generation datasets, achieving superior performance across the board.

\end{itemize}

\begin{figure*}[t]
  \centering
\includegraphics[width=\textwidth]{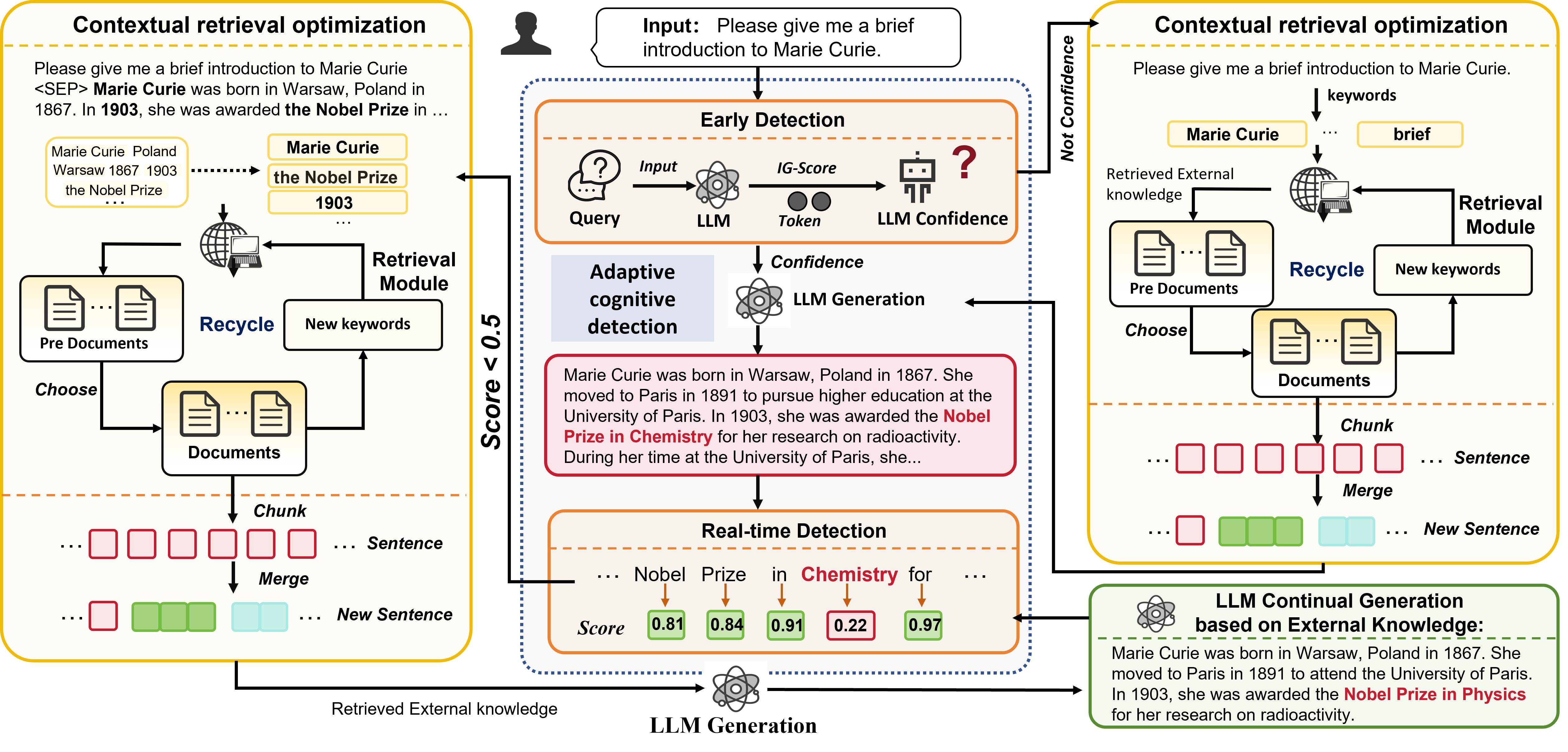} 
     \vspace{-15px}
  \caption {Detailed technical framework of DioR. It is mainly divided into two technical aspects: 1) Adaptive cognitive detection; and 2) Contextual retrieval optimization.}
  \label{fig:DioR}
\end{figure*}

\section{Related Work}

Large language models (LLMs) have shown outstanding performance in solving downstream tasks \cite{kumar2024large}; however, due to their inability to integrate new information, they are prone to generating hallucinations during generation \cite{su2024unsupervised, ramprasad2024analyzing}. To address this issue, Retrieval-Augmented Generation (RAG) is an effective strategy, enhancing the model's ability to leverage non-parametric knowledge by retrieving external data resources \cite{jiang2022retrieval, jiang2024longrag}.

The initial RAG paradigm primarily focused on single-turn retrievals, such as Single-round RAG (SR-RAG) methods like KNN-LM \cite{wang2023knn}, ReAtt \cite{jiang2022retrieval}, REPLUG \cite{shi2023replug}, and UniWeb \cite{li2023web}, which retrieve relevant passages from external corpora based on the initial query and append them to the LLM’s input. Single-turn retrieval indeed shows significant improvements in the context of simple tasks. However, when confronted with more complex tasks, such as multi-step reasoning, long-form question answering, and chain-of-thought reasoning \cite{bechard2024reducing, DBLP:conf/acl/SuTA0024}, relying solely on the initial input for RAG often fails to provide sufficient effective external knowledge, resulting in limited improvements in addressing hallucinations \cite{wu2024cotkr}.

To address this issue, researchers have explored multi-turn retrieval-augmented strategies (Dynamic RAG). Fixed Length RAG (FL-RAG) methods, such as RETRO \cite{borgeaud2022improving} and ICRALM \cite{ram2023context}, trigger the retrieval module after every n token, using the generated tokens from the previous token window as the query. Fixed Sentence RAG (FS-RAG) methods, such as IRCot \cite{Trivedi2022}, trigger retrieval after each sentence, treating each generated sentence as a new query. However, these multi-turn retrieval approaches do not fully consider the real-time needs of the LLM during the generation process, which may lead to suboptimal results. In contrast, FLARE \cite{jiang2023active} triggers retrieval when uncertain tokens are encountered (i.e., when the probability of any token in the text falls below a certain threshold), and inspired by this, Dragin \cite{DBLP:conf/acl/SuTA0024}, a novel dynamic RAG method, determines the timing and content of retrieval based on the LLM’s attention entropy and token relevance.

However, current dynamic RAG methods fail to predict whether the LLM has the capability to answer a question prior to generation, thereby triggering retrieval in advance. Moreover, most methods often rely on static rules, leading to ineffective timing for retrieval triggers during the generation process. Additionally, existing dynamic RAG approaches lack optimization strategies for the document retrieval process and the quality of retrieved documents, such as relevance and length, which significantly hampers the LLM’s performance


\section{Proposed Method: DioR} 
In this section, we present the \textbf{DioR}, an innovative dynamic Retrieval-Augmented Generation (RAG) method, as illustrated in Figure \ref{fig:DioR}. DioR consists of two main components: \textit{adaptive cognitive detection} and \textit{contextual retrieval optimization}.


\subsection{Adaptive cognitive detection}

In view of the limitation of existing dynamic RAG methods that \textbf{\textit{Lack an effective mechanism to control retrieval triggers}}, we propose an innovative approach, \textbf{Adaptive cognitive detection}, which aims to effectively judge the optimal \textit{timing} of retrieval based on the cognitive characteristics of large language models (LLMs). We specifically explain the method from two perspectives.

\subsubsection{Is LLM confident in answering this?}

We mentioned before that existing dynamic RAG methods all determine whether hallucinations occur after LLM generates content. However, if an LLM exhibits a lack of sufficient confidence when confronted with a particular question before generating, does it mean the model is more likely to generate incorrect information in uncertain situations, potentially leading to hallucinations? Therefore, can we intervene in the model's output too early, before it generates such hallucinations?

\begin{algorithm}[H]
\caption{Early Detection Dataset Construction and Training}
\begin{algorithmic}[1]
\State \textbf{Input:} Wikipedia 

\State \textbf{Output:} Trained RNN classifier \(RNN_C\)

\State \textbf{Step 1: Data Preparation}
\For{each question-answer pair \(Q, R\)}
    \State Generate answer \(A\) using LLaMA2-7B based on question \(Q\)
    \If{\(A\) matches \(R\) or contains \(R\) (i.e., \(R \subseteq A\))}
        \State Label as Non-Hallucination
    \Else
        \State Label as Hallucination
    \EndIf
    \State Extract IG Attribution Entropy (IG features) from question \(Q\)
    \State Store \((Q, A, R, \text{IG}, \text{Label})\)
\EndFor

\State \textbf{Step 2: Model Training}
\State Initialize RNN-based classifier \(RNN_C\)
\For{each data point \((Q, A, R, \text{IG}, \text{Label})\)}
    \State Feed \((Q, A, R, \text{IG})\) into the RNN
    \State Compute predicted label \(\hat{y}\) and loss \(\mathcal{L}(\hat{y}, \text{Label})\)
    \State Backpropagate and update model parameters to minimize \(\mathcal{L}\)
\EndFor

\State \textbf{Step 3: Output} Trained classifier \(RNN_C\)
\end{algorithmic}
\label{al:1}
\end{algorithm}

Therefore, we plan to train a detector to determine whether early intervention in retrieval \textbf{timing} is necessary to optimize the subsequent generation process. Specifically, \textbf{Early Detection} aims to assess the LLM's confidence level regarding a given question before generation, and based on this assessment, decide whether retrieval should occur prior to generation. The specific dataset construction and training process is detailed in Algorithm \ref{al:1} and Appendix \ref{sec:appendix1.1} (Algorithm Explanation).

Based on this, we trained an RNN classifier on a large Wikipedia dataset to determine whether to trigger RAG based on the LLM's confidence level in responding to a given question.

\subsubsection{Does LLM have accurate output?}

\begin{algorithm}[H]
\caption{Real-time Detection Dataset Construction and Training}
\begin{algorithmic}[1]
\State \textbf{Input:} Wikipedia \(W = \{w_1, w_2, ..., w_n\}\)
\State \textbf{Output:} Trained MLP classifier \(MLP_C\)

\State \textbf{Step 1: Data Preparation}
\For{each article \(w_i \in W\)}
    \State Truncate article based on entity positions
    \State Generate text \(t_i\) using LLaMA2-7B from truncated article
\EndFor

\State \textbf{Step 2: Entity Extraction and Comparison}
    \State Extract entities \(e_i\) from original article \(w_i\)
    \State Extract entities \(n_i\) from generated text \(t_i\)
    \For{each entity \(n_j \in n_i\), \(e_j \in e_i\)}
        \State Compare \(e_j\) and \(n_j\) for cosine similarity
        \If{similarity between \(e_j\) and \(n_j\) is high}
            \State Consider \(e_j\) and \(n_j\) as the same entity (no hallucination)
        \Else
            \State Mark \(t_i\) as hallucination (0)
        \EndIf
    \EndFor
    
    \State \textbf{Step 3: Dataset Construction}
    \State Construct data point \(D_i = \{w_i, e_i, t_i, n_i, H_i\}\), where \(H_i\) is hallucination label (0 for hallucination, 1 for non-hallucination)
    \State Store data point \(D_i\) in dataset \(D_n\)

\State \textbf{Step 4: Model Training}
\State Train MLP classifier \(MLP_C\) on dataset \(D_n\)
\For{each data point \((w_i, e_i, t_i, n_i, H_i) \in D_n\)}
    \State Feed \(w_i, e_i, t_i, n_i\) into the MLP classifier
    \State Compute predicted hallucination label \(\hat{H_i}\) and loss \(\mathcal{L}(\hat{H_i}, H_i)\)
    \State Backpropagate and update classifier parameters to minimize \(\mathcal{L}\)
\EndFor

\State \textbf{Step 5: Output} Trained classifier \(MLP_C\)
\end{algorithmic}
\label{al:2}
\end{algorithm}

As mentioned earlier, most existing dynamic RAG frameworks rely on static, predefined rules to trigger the retrieval module. For example, retrieval is initiated when the generation probability of a token during the LLM's output process falls below a predefined threshold, treating such low-probability tokens as potential hallucinations.

However, hallucination detection should not rely on generation probability. While the generation probability could indicate model confidence \cite{farquhar2024detecting, zhang2024luq}, it does not directly reflect the accuracy of the LLM-generated text, and therefore may not be a reliable criterion for the occurrence of hallucinations.

Thus, the \textbf{Real-time Detection} aims to monitor the generation process of the LLM and detect in real time whether the output tokens from LLMs are likely to be hallucinations, as the standard of the retrieval timing. The specific dataset construction and training process is described in Algorithm \ref{al:2} and Appendix \ref{sec:appendix1.2} in detail.

Therefore, we have trained an MLP-based hallucination detector on a large corpus of Wikipedia data to ensure its effectiveness in accurately identifying hallucinations.

\subsection{When to retrieve is needed?}
\label{sec:3.2}

The appropriate retrieval \textit{timing} helps LLM to avoid hallucinations in the generation process with maximum efficiency. So we define the detection process of the above two classifiers as follows:

\paragraph{Definition 1.}
\textit{Let \(C(Q)\) denote the LLM's confidence in question \(Q\). \(IG(Q)\) represents the Attribution Entropy (IG features) (more ref. \ref{sec:proof}).}
\begin{equation}
    IG(Q) = \sum_{j=1}^{N} \frac{-\text{IG}_j}{\sum_{k=1}^{N} \text{IG}_k} \log \left( \frac{\sum_{k=1}^{N} \text{IG}_k}{\text{IG}_j} \right).
\end{equation}
\begin{equation}
    C(Q) = \mathbb{I}(\text{Softmax}(f_{\text{RNN}}(\text{IG}(Q))) > 0.5).
\end{equation}
\textit{If \(C(Q)\) is 1, the retrieval trigger. We calculate the IG attribution value \(IG_{i}\) separately for each token \(t_i\) of \(Q\), and the mean of all token is \(IG_{mean}\). For selecting candidate keywords for retrieval (the model tends to focus on), we following bellow:}
\begin{equation}
     \text{\textit{If} }  IG_{i} > IG_{mean} \quad \Rightarrow \quad t_i \text{\textit{ as candidate.}}
\end{equation}

\paragraph{Definition 2.}
\textit{Let \(t_j\) as the generating token of LLMs. \(P_{t_j}\) as hallucination probability of \(t_j\).}
\begin{equation}
    P_{t_j} = \sigma(f_{MLP}(t_j)).
\end{equation}
\textit{If \(P_{t_j}\) (the sigmoid layer score \(\sigma\)) is below 0.5, the token \(t_j\) is flagged as a hallucination and the retrieval trigger. Then, use spaCy to extract all entities from the generated text, convert them into tokens, filter out the tokens labeled as hallucinations, and ultimately attain valid candidate keywords.}




\subsection{Contextual retrieval optimization}

Once the LLM is determined to lack confidence in answering a question or when hallucinations are detected, triggering RAG becomes necessary. The next step in the RAG framework is to generate queries and retrieve relevant information from external databases to support the LLM in continuing text generation.  However, existing dynamic RAG methods suffer from the limitation of \textbf{\textit{Lack of effective scrutiny of retrieval content}}.



To address the limitation, we propose a novel method called \textbf{Contextual retrieval optimization}, consisting of two key steps: pre-retrieval and post-retrieval. Specifically, as follows.

\subsubsection{Pre-retrieval}

Existing dynamic RAG frameworks rely solely on the token confidence scores (e.g., attention weights) of the most recent sentences generated by the LLM to select the final retrieval keyword. This approach, which selects keywords based on recent sentences or token confidence, struggles to accurately capture subtle semantic relationships between contextual information, potentially leading to suboptimal retrieval results. Therefore, we performed the following optimizations in the selection of keywords.

We analyzed the entire text generated by the LLM using the two detectors (sec. \ref{sec:3.2}, Early and Real-time Detection) and identified a set of candidate keywords. Then, we assess the importance of each candidate keyword (token) through four evaluation metrics. This helps the model choose the most important keyword (token) to retrieve.

We first compute the attention scores of each token using a multi-head attention mechanism. By representing attention across different heads, the model can focus on various semantic and syntactic features of the text, offering a more comprehensive understanding of which parts of the text are most important for retrieving external information. Let \( A_i \) represent the attention score for token \( i \).
Next, we incorporate TF-IDF scores to evaluate the information density of each token. We assign higher priority to tokens with greater information density. We then calculate a positional score based on the relative position of each token in the text, taking into account its contextual role in text generation. The positional score for token \( i \) is computed as: $P_i = \frac{\mathrm{Pos}(i)}{N}$, where $\mathrm{Pos}(i)$ and $N$ are the position of the token and the total number of tokens respectively. 
Finally, we assess the semantic relevance between each token and the query term by calculating the cosine similarity \( S_i \) between the word embedding vectors of the token \( i \) and the query \( \vec{q} \).

The overall importance score \( I_i \) for each token is computed as a weighted sum of four components, which enables more accurate identification of relevant tokens and enhances subsequent retrieval.

DioR ranks tokens in descending order of importance based on \( I_i \), it prioritizes those most relevant to the retrieval task. These top-ranked tokens facilitate the extraction of the most relevant external knowledge with the question. Advanced retrieval models, such as BM25, are then used to retrieve relevant documents from an external knowledge base, enhancing the precision of the retrieval process. Then, we put the retrieved documents in the candidate area for the next step of optimization.

\subsubsection{Post-retrieval}
\label{sec:3.3.2}

After retrieving relevant documents from the knowledge base based on token priority, we need to design a dynamic optimization strategy to enhance the accuracy and relevance of documents for real-time retrieval needs. Specifically, since current dynamic RAG frameworks retrieve all documents in a single batch per query trigger, lacking flexibility and failing to fully understand the task context, we discard this strategy in favor of stepwise retrieval.

For example, if five documents need to be retrieved, we first choose the top two most relevant from the candidate area (n/2, where n is the number of remaining documents to be retrieved). Next, we perform keyword extraction on these two documents, focusing on newly emerging keywords or concepts. The newly emerged concepts are merged with the original keywords, and a new round of retrieval is conducted based on the updated keyword set. This process is repeated until the required number of documents is selected, providing accurate information support for subsequent generation.


Once the document selection is complete, the next step is to input the selected documents into the LLM. If any individual document is too long, it may hinder the model's understanding and reasoning. To address this, we design a sentence-level segmentation approach for the document. However, simple sentence segmentation may lead to semantic breaks, potentially causing misinterpretations by the model. To solve this problem, we reassemble the sentences into shorter, semantically coherent segments, ensuring that the split content maintains logical continuity.
Specifically, for each sentence \( x \), we first split it into sub-clauses \( x_1, x_2, \dots, x_n \), then evaluate the scores for various sub-clause combinations. For instance, we compare the score of the combination of \( x_1 \) and \( x_2 \) with that of \( x_1 \) alone. If the combination score improves, we proceed by adding \( x_3 \), comparing the score of \( x_1, x_2, x_3 \) with that of \( x_1, x_2 \), and repeat this process until the score decreases. At that point, the sub-clauses are grouped into a block. This method enables us to break each document \( d_i \) into several semantically coherent and shorter blocks (e.g., \( d_{i1}, d_{i2}, \dots, d_{ik} \)), mitigating the negative impact of long texts on the model's comprehension and optimizing its reasoning performance. This ensures that each input is semantically clear and concise.

We integrate all the segmented blocks from the retrieved documents into the LLM prompt template, please refer to the following for details:

\text{External Knowledge After Chunk: } 

[1] \, $d_1[(1). d_{11}, (2). d_{12}, \dots, (u).d_{1u}]$

[2] \, $d_2[(1). d_{21}, (2). d_{22}, \dots, (t).d_{2t}]$

...
 
[3] \, $d_i[(1). d_{i1}, (2). d_{i2}, \dots, (k). d_{ik}]$
 \dots

Using the external knowledge provided above, please answer the following question:

Question: [Ques.] 

Answer: 
Insert truncated output [ ] and additional relevant details here.

This integration method effectively addresses the knowledge gaps during the LLM generation process. Specifically, when hallucinations occur in the content generated by the LLM, we make it a truncation point, and we introduce externally retrieved knowledge to allow the LLM to continue generating more accurate and comprehensive content from the truncation point. This enhances the model's understanding of the documents and ensures that the generated content is more precise and relevant.

\section{Experiment}

\subsection{Experimental Setups}

The comprehensive and detailed descriptions of the experimental setups, including the \textit{datasets}, \textit{evaluation metrics}, and \textit{implementation specifics}, can be found in Appendix \ref{sec:appendix2}.

\begin{table*}[h]
\centering
\caption{Comparison of EM, F1, Precision, and Recall scores between the Base method and DioR across various datasets and retrieval strategies. The highest scores for each metric in each dataset are highlighted in underline.}
\scalebox{1}{
\begin{tabular}{l|ccccccccccccc}
\toprule
\multirow{2}{*}{\textbf{Matrix}} & \multicolumn{2}{c}{\textbf{2WikiMultihopQA}} & \multicolumn{2}{c}{\textbf{HotpotQA}} & \multicolumn{2}{c}{\textbf{IIRC}} & \multicolumn{2}{c}{\textbf{StrategyQA}} \\
\cmidrule(r){2-3} \cmidrule(r){4-5} \cmidrule(r){6-7} \cmidrule(r){8-9}
 & Base & \textbf{DioR} & Base & \textbf{DioR} & Base & \textbf{DioR} & Base & \textbf{DioR} \\
\midrule
EM (BM25) & 0.214 & \textbf{0.254} & 0.219 & \underline{\textbf{0.274}} & 0.156 & \underline{\textbf{0.201}} & 0.639 & \underline{\textbf{0.659}} \\
EM (SGPT) & 0.209 & \textbf{0.226} & 0.202 & \textbf{0.212} & 0.125 & \textbf{0.178} & 0.604 & \textbf{0.616} \\
EM (SBERT) & 0.231 & \underline{\textbf{0.266}} & 0.165 & \textbf{0.205} & 0.142 & \textbf{0.153} & 0.645 & \textbf{0.654} \\
\midrule
F1 (BM25) & 0.282 & \underline{\textbf{0.335}} & 0.314 & \underline{\textbf{0.379}} & 0.188 & \underline{\textbf{0.245}} & 0.639 & \underline{\textbf{0.659}} \\
F1 (SGPT) & 0.278 & \textbf{0.292} & 0.301 & \textbf{0.307} & 0.153 & \textbf{0.224} & 0.604 & \textbf{0.616} \\
F1 (SBERT) & 0.294 & \textbf{0.328} & 0.244 & \textbf{0.303} & 0.172 & \textbf{0.179} & 0.645 & \textbf{0.654} \\
\midrule
Pre. (BM25) & 0.288 & \underline{\textbf{0.342}} & 0.331 & \underline{\textbf{0.399}} & 0.195 & \underline{\textbf{0.255}} & 0.639 & \underline{\textbf{0.659}} \\
Pre. (SGPT) & 0.284 & \textbf{0.298} & 0.319 & \textbf{0.333} & 0.159 & \textbf{0.229} & 0.604 & \textbf{0.616} \\
Pre. (SBERT) & 0.298 & \textbf{0.334} & 0.256 & \textbf{0.324} & 0.176 & \textbf{0.185} & 0.645 & \textbf{0.654} \\
\midrule
Rec. (BM25) & 0.285 & \underline{\textbf{0.345}} & 0.316 & \underline{\textbf{0.381}} & 0.196 & \underline{\textbf{0.243}} & 0.639 & \underline{\textbf{0.659}} \\
Rec. (SGPT) & 0.281 & \textbf{0.299} & 0.305 & \textbf{0.312} & 0.156 & \textbf{0.219} & 0.604 & \textbf{0.616} \\
Rec. (SBERT) & 0.299 & \textbf{0.332} & 0.255 & \textbf{0.303} & 0.180 & \textbf{0.189} & 0.645 & \textbf{0.654} \\ 
\bottomrule
\end{tabular}
}
\label{tab:all}
\end{table*}

\begin{table*}[h]
\centering
\caption{Comparison of DioR and other RAG methods for LLaMA2-7B-CHAT across four different datasets}
\setlength{\tabcolsep}{3pt} 
\resizebox{\textwidth}{!}{%
\begin{tabular}{l|c|c|ccccccccc}
\toprule
\textbf{Dataset} & \textbf{Metric} & \textbf{DioR} &  \textbf{SEAKR} &\textbf{RaDIO} & \textbf{Dragin} & \textbf{wo-RAG} & \textbf{SR-RAG} & \textbf{FL-RAG} & \textbf{FS-RAG} & \textbf{FLARE} \\
\midrule
\multirow{2}{*}{\textbf{2WikiMultihopQA}} 
& EM & \textbf{0.266} & 0.264 & 0.254 & 0.231 & 0.146 & 0.169 & 0.112 & 0.189 & 0.143 \\
& F1 & \textbf{0.335} & 0.330 & 0.317 & 0.294 & 0.223 & 0.255 & 0.192 & 0.265 & 0.213 \\
\midrule
\multirow{2}{*}{\textbf{HotpotQA}} 
& EM & \textbf{0.274} & 0.261 & 0.246 & 0.219 & 0.184 & 0.164 & 0.146 & 0.214 & 0.149 \\
& F1 & \textbf{0.379} & 0.365 & 0.351 & 0.314 & 0.275 & 0.150 & 0.211 & 0.304 & 0.221 \\
\midrule
\multirow{2}{*}{\textbf{IIRC}} 
& EM & \textbf{0.201} & 0.195 & 0.196 & 0.156 & 0.139 & 0.187 & 0.172 & 0.178 & 0.136 \\
& F1 & \textbf{0.245} & 0.235 & 0.239 & 0.188 & 0.173 & 0.226 & 0.203 & 0.216 & 0.164 \\
\midrule
\multirow{1}{*}{\textbf{StrategyQA}} 
& Pre. & \textbf{0.659} & 0.650 & 0.654 & 0.645 & 0.659 & 0.645 & 0.634 & 0.629 & 0.627 \\
\bottomrule
\end{tabular}
}
\label{tab:rag}
\end{table*}

\subsection{Experimental Results}
In this subsection, we mainly discuss the most important experimental results, which aim to comprehensively evaluate DioR's performance across four datasets with various competitors. The additional experimental results, which provide further insights into the performance of our method, are discussed in greater detail in Appendix \ref{sec:appendix3}.

\begin{table*}[h]
\centering
\caption{Ablation experiments. ED: Early-Detection, RD: Real-time Detection, Pre-R/Post -R: Pre/Post -retrieve.}
\scalebox{1}{
\begin{tabular}{l|c|c|cccc}
\toprule
\textbf{Dataset} & \textbf{Metric} & \textbf{DioR} & w/o ED & w/o RD & w/o Pre-R & w/o Post-R \\
\midrule
\multirow{2}{*}{\textbf{2WikiMultihopQA}} 
& EM & \textbf{0.266} & 0.258 & 0.239 & 0.249 & 0.260  \\
& F1 & \textbf{0.335} & 0.327 & 0.301 & 0.306 & 0.322  \\
\midrule
\multirow{2}{*}{\textbf{HotpotQA}} 
& EM & \textbf{0.274} & 0.266 & 0.223 & 0.237 & 0.249  \\
& F1 & \textbf{0.379} & 0.363 & 0.319 & 0.334 & 0.356  \\
\midrule
\multirow{2}{*}{\textbf{IIRC}} 
& EM & \textbf{0.201} & 0.191 & 0.169 & 0.172 & 0.197  \\
& F1 & \textbf{0.245} & 0.233 & 0.197 & 0.206 & 0.240  \\
\midrule
\multirow{1}{*}{\textbf{StrategyQA}} 
& Pre. & \textbf{0.659} & 0.652 & 0.646 & 0.648 & 0.655  \\
\bottomrule
\end{tabular}
}
\label{tab:aba}
\end{table*}

\subsubsection{Overall Results}
In this part, we present results comparing the performance of the Base (Dragin \cite{DBLP:conf/acl/SuTA0024}) and DioR (Using LLaMA2-7B). As shown in Table \ref{tab:all} in detail.

In 2WikiMultihopQA, the DioR method consistently outperforms the Base method across all retrieval methods and evaluation metrics. Specifically, DioR achieves a notable increase in EM, with a score of 0.254 (BM25) compared to the Base's 0.214. Similarly, F1, Precision, and Recall scores show advantages for DioR, surpassing the Base.

In HotpotQA. For EM, DioR reaches 0.274 (BM25), an improvement over the Base's 0.219. The F1 score is also better for DioR, achieving 0.379 (BM25) versus the Base's 0.314. Pre. and Rec. follow a similar trend. It is worth noting that under the SBERT retrieval strategy, the Pre. is improved by 0.068, the largest increase compared with BM25 and SGPT.

In the IIRC dataset, DioR demonstrates a consistent advantage across metrics. The EM score increases from 0.156 (Base) to 0.201 (DioR, BM25). F1, Pre., and Rec. scores also show significant improvement, with DioR reaching a Pre. of 0.255 (BM25) and Rec. of 0.243 (BM25), higher than the Base method by a noticeable margin. 

In the StrategyQA dataset, DioR outperforms the Base method across all retrieval methods. The EM score increases from 0.639 (Base) to 0.659 (DioR, BM25). F1, Pre., and Rec. show similar improvements, with DioR achieving a Precision score of 0.659 and a Recall score of 0.659 (both BM25), marking a consistent performance boost.

Overall, results show that DioR outperforms the Base method across all datasets and evaluation metrics, providing a more robust solution for question-answering tasks and effectively addressing hallucinations while enhancing LLM reasoning.

\subsubsection{Comparison with Other RAG methods}

In this part, we present a comparison of the performance of DioR and various RAG-based methods (Using LLaMA2-7B). As shown in Table \ref{tab:rag}.

On the 2WikiMultihopQA dataset, DioR achieves the highest performance. Specifically, DioR scores 0.266 for EM, surpassing some popular dynamic RAG methods, such as SEAKR, RaDIO, and Dragin. Similarly, DioR leads in F1, with a score of 0.335. Other RAG methods, such as wo-RAG (EM: 0.146, F1: 0.223) and SR-RAG (EM: 0.169, F1: 0.255), perform notably worse.

In HotpotQA, DioR also outperforms other methods, achieving higher EM (0.274) and F1 (0.379) scores than popular dynamic RAG methods such as RaDIO (EM: 0.246, F1: 0.351), SEAKR (EM: 0.261, F1: 0.365), and Dragin (EM: 0.231, F1: 0.294). In contrast, wo-RAG and SR-RAG show significantly lower scores (EM: 0.184, F1: 0.275 for wo-RAG; EM: 0.164, F1: 0.150 for SR-RAG). 

In the IIRC dataset, DioR achieves the highest EM (0.201) and F1 (0.245) scores, outperforming other RAG methods such as Dragin (EM: 0.156, F1: 0.188), FL-RAG (EM: 0.172, F1: 0.203) and FS-RAG (EM: 0.178, F1: 0.216). 

On the StrategyQA dataset, DioR achieves the highest Precision score (0.659), surpassing the RaDIO (0.654), Dragin (0.645), SEAKR (0.650), and other RAG methods. While wo-RAG shows similar performance (0.659) due to the dataset is not complex, DioR still leads in all datasets, which is indicative of its overall more reliable performance.

Thus, the experimental results demonstrate that DioR offers a robust solution for handling LLM-generated hallucinations, making it a superior choice for tasks requiring high factual accuracy, such as those involving complex, multi-hop reasoning or multi-turn question answering.

\subsubsection{Ablation Experiment}

In this part, we mainly test the effectiveness of the following two components of our proposed method: adaptive cognitive detection and contextual retrieval optimization. As shown in Tabel. \ref{tab:aba}.

It is worth noting that if we remove the Real-time Detection module, retrieval is triggered directly when the generation probability of a token is below 0.5. If we remove the ``Pre-retrieval'' step, retrieval is performed sequentially based on the order of token appearances. The remaining two modules can be directly removed: removing Early Detection eliminates the need to assess the LLM's ability to answer in advance while removing ``Post-retrieval'' leads to single-batch retrieval, where all documents are included in the prompt input to the LLM (the prompt for LLMs reference Appendix \ref{sec:appendix3.4}).

Based on the results of the ablation experiments, we can observe the effectiveness of each component in enhancing DioR across four datasets. This further validates the efficacy of our proposed method in addressing the two limitations of existing dynamic RAG approaches.

\section{Conclusion and Future work}

In this paper, we investigate the effectiveness of Retrieval-Augmented Generation (RAG) techniques in mitigating hallucination issues in large language models (LLMs). However, existing dynamic RAG methods face significant limitations in two key aspects, \textbf{\textit{1) Lack of an effective mechanism to control retrieval triggers}}, and \textbf{\textit{2) Lack of effective scrutiny of retrieval content}}. To address these limitations, we propose an innovative dynamic RAG approach, \textbf{DioR} (Adaptive Cognitive \textbf{D}etect\textbf{i}on and C\textbf{o}ntextual \textbf{R}etrieval Optimization), achieving when retrieval is needed and what to retrieve for LLMs is useful. Compared to existing popular dynamic RAG methods, DioR demonstrates superior performance across four knowledge-intensive generation datasets, proving the effectiveness of our method in improving.

Looking ahead, we will refine DioR, especially in its ability to tackle complex problems in a step-by-step manner. By integrating more refined reasoning strategies, we aim to further elevate the model's performance on intricate tasks.

\section*{Limitations}
We acknowledge the limitations of this paper. Specifically, we face the performance impact of retrieving long documents for model inference. While we have implemented chunking to shorten the length of individual knowledge pieces and improve the model’s understanding of each, the total length of all input knowledge remains unchanged. In the future, we could introduce a model that summarizes the key points of each document, reducing the overall length and focusing on the core content for retrieval, thereby improving both the efficiency and the relevance of the retrieved knowledge.

On the other hand, we have identified that complex problems often pose significant challenges when approached through a single, direct reasoning process. For instance, in mathematical problems, attempting to solve them in one step can lead to increased computational complexity, higher error rates, and difficulty in identifying intermediate issues. To address these challenges, we plan to adopt a step-by-step reasoning approach in the future. Specifically, we will plan to break down complex problems into a series of smaller, more manageable sub-problems. By tackling each sub-problem sequentially, we can enhance the clarity and accuracy of the reasoning process, making it easier to identify and correct errors at each stage, and ultimately improve the overall effectiveness and reliability of large language models' reasoning.

\section*{Acknowledgment}
This work was supported by the National Key R\&D Program of China under Grant No. 2022YFC3303600, the Zhejiang Provincial Natural Science Foundation of China under Grant No. LY23F020010, and the National Natural Science Foundation of China under Grant Nos. 62337001.

\bibliography{custom}

\newpage
\appendix

\section{Adaptive cognitive detection algorithm details}
\label{sec:appendix1}

\begin{figure*}[t]
  \centering
  \includegraphics[width=\textwidth]{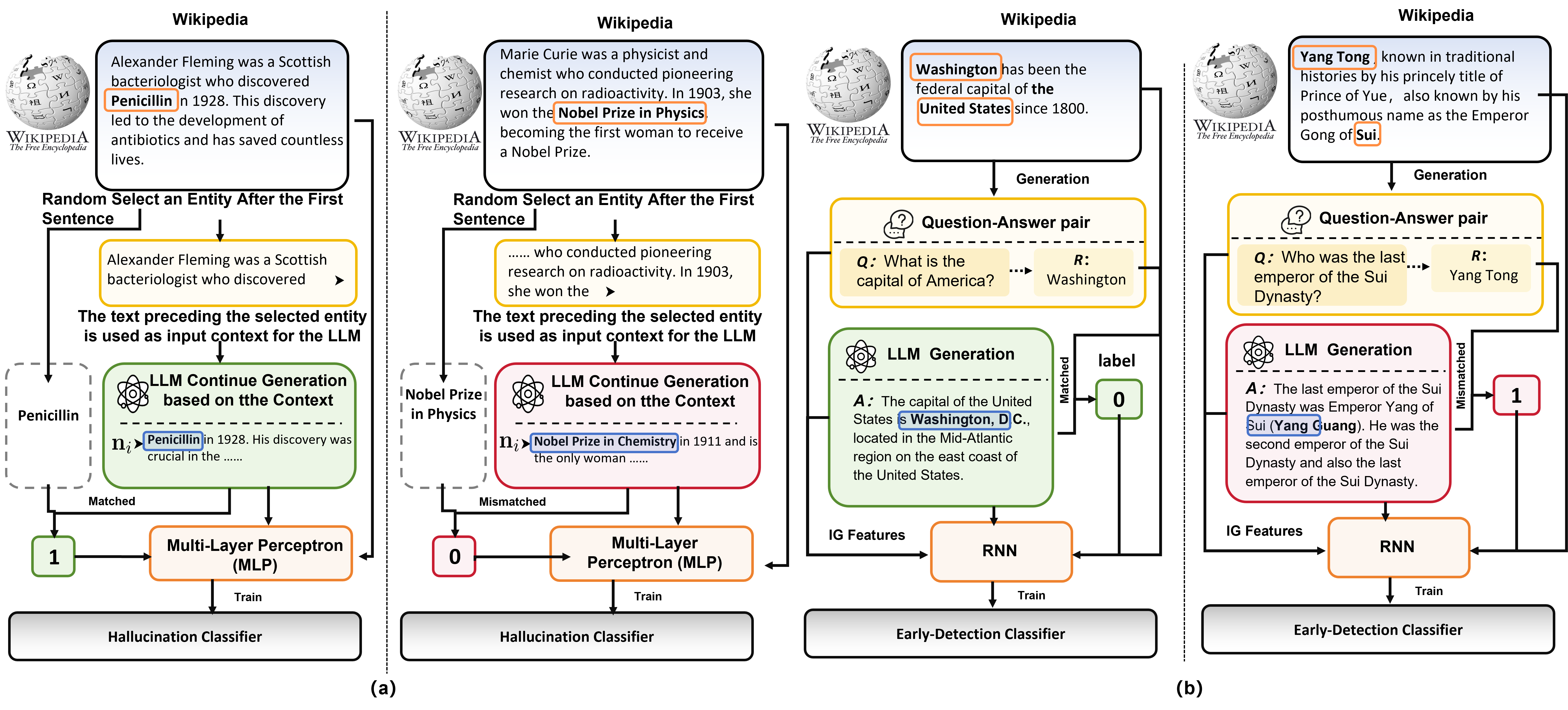} 
  \caption {The construction process of Real-time Detection (a) and Early Detection (b) datasets}
  \label{fig:detec}
\end{figure*}

In Section 3, we build two hallucination detection classifiers, as shown in Figure \ref{fig:detec}. We describe the algorithm described above in detail as follows and the effectiveness analysis of these two classifiers is shown in the Appendix \ref{sec:appendix3.3}:

\subsection{Early Detection}
\label{sec:appendix1.1}

We choose the Wikipedia dataset and used LLaMA2-7B-CHAT for generative question answering. For a randomly selected question \( Q \), we generate an answer \( A \) with the model, where each \( Q \) contains a ground truth answer \( R \), as described in lines 3 to 5 of Algorithm \ref{al:1}. We consider that if the generated answer \( A \) is inconsistent with the factual details of \( R \), then it is factually incorrect, and the model's response is a hallucination \cite{jiang2024survey, ji2023survey}. This definition aligns with previous generative question-answering studies, which view factually incorrect statements as hallucinations. However, since LLM outputs can be quite long, merely matching the generated answer \( A \) with the reference answer \( R \) via exact string comparison is insufficient for judging the correctness of \( A \). For example, assume \( Q = \) [Where is the capital of America?] and \( R = \) [Washington]. Whether LLaMA generates the answer \( A \) as "Washington" or "Washington D.C." both represent correct answers. Therefore, we consider that the reference answer \( R \) is contained within the generated answer \( A \), i.e., \( R \subseteq A \), and the answer is correct, as described in lines 6 to 10 of Algorithm \ref{al:1}.

To effectively determine whether the LLM has enough information to answer a question, we can assess the model’s focus on certain words within the question, i.e., whether the LLM can capture the key points of the question. We use the Integrated Gradients (IG) method to generate feature attribution, which indicates the importance of each feature in the specific prediction \cite{snyder2024early}. This method is typically used to inspect the behavior of the model and uncover potential problematic patterns. In simple terms, if we calculate the feature attribution based on the LLM's attention, if the LLM’s output does not generate hallucinations, i.e., if it focuses on certain keywords, the attribution entropy will be low (focusing on keywords, keywords' attribution is high). Otherwise, when hallucinations are generated, the model will focus on more input words, resulting in higher attribution entropy (the focus is more dispersed, and all word attribution is relatively even).  Specific proof of attribution entropy process reference \ref{sec:proof}.

Thus, we construct an Early-Detection dataset consisting of \( Q, A, R, \) and the IG-generated feature attribution values, along with hallucination labels, as described in lines 11 and 12 of Algorithm \ref{al:1}.

Due to the sequential nature of the data, we use a Recurrent Neural Network (RNN) for training, as RNNs are adept at handling sequential data and effectively capture contextual dependencies within the input sequence, as described in lines 14 to 21 of Algorithm \ref{al:1}. During training, we use the standard binary cross-entropy loss function to predict the \( \hat{y}_i \) given \( Q, A, R, \) and \( IG \), as follows:

\begin{equation}
    L(\theta) = - \sum_{i=1}^{N} \left[ y_i \log(\hat{y}_i) + (1 - y_i) \log(1 - \hat{y}_i) \right], 
\end{equation}
where $\hat{y}_i = \text{sigmoid}\left(f(Q_i, A_i, R_i, IG_i, \theta)\right)$ and $f(Q_i, A_i, R_i, IG_i, \theta)$ is the model's output, calculated from the model parameters $\theta$ and the inputs $Q_i, A_i, R_i, IG_i$.

\subsubsection{Proof of Attribution Entropy Behavior}
\label{sec:proof}

Uniform distribution of attribution values: If the model pays almost equal attention to each input feature (for example, the attribution value of each feature is similar), the entropy value will be high because the model as a whole does not tend to favor a specific feature, but pays attention to multiple features. In this case, the system is more uncertain or chaotic, so the attribution entropy is large.

Concentrated attribution values: If the model's attention is mainly focused on a few features, while the attribution values of other features are small or close to zero (for example, the attribution values of some features are significantly higher than other features), the entropy value will be small because the model's attention is not scattered, but concentrated on a few features, which reduces the uncertainty of the system, so the attribution entropy is small.

The following proves the attribution entropy, specifically:

\begin{enumerate}[leftmargin=*]
    \item \textbf{Definition of Attribution Entropy:}

The attribution entropy \( H(P) \) of a probability distribution \( P = \{p_1, p_2, ..., p_n\} \) is defined as:

\begin{equation}
    H(P) = - \sum_{i=1}^{n} p_i \log p_i,
\end{equation}
where \( \sum_{i=1}^{n} p_i = 1 \).

\item \textbf{Uniform Distribution (Hallucination):}

For a uniform distribution, where each probability \( p_i = \frac{1}{n} \), the attribution entropy becomes:

\begin{equation}
    H(P) = - \sum_{i=1}^{n} \frac{1}{n} \log \frac{1}{n} = \log n
\end{equation}

\item \textbf{Concentrated Distribution (Non-Hallucination):}

Now, consider a distribution where most of the probability mass is concentrated on one element. For example, let \( p_1 = 1 - \epsilon \) and \( p_2 = p_3 = \dots = p_n = \frac{\epsilon}{n-1} \), where \( \epsilon \to 0 \). The attribution entropy is:

\begin{equation}
    \begin{split}
        H(P) &= - \left( (1 - \epsilon) \log (1 - \epsilon) \right) \\
        &\quad - (n-1) \times \frac{\epsilon}{n-1} \log \frac{\epsilon}{n-1}
    \end{split}
\end{equation}

When \( \epsilon \) is small, we can approximate \( \log(1 - \epsilon) \approx -\epsilon \). Thus:

\begin{equation}
    \begin{split}
        H(P) &\approx -\left( (1 - \epsilon)(-\epsilon) \right) \\
        &\quad - (n - 1) \times \frac{\epsilon}{n - 1} \log \frac{\epsilon}{n - 1}
    \end{split}
\end{equation}

This simplifies to:

\begin{equation}
    H(P) \approx \epsilon- \epsilon^2 - \epsilon \log \frac{\epsilon}{n - 1}
\end{equation}

As \( \epsilon \to 0 \), \( H(P) \to 0 \), indicating that the attribution entropy is minimized for a concentrated distribution.

\end{enumerate}

\subsection{Real-Time Detection}
\label{sec:appendix1.2}

Specifically, 
we consider that the part of the LLM output most prone to hallucinations typically involves changes in entities (e.g., incorrect dates or names) \cite{huang2024survey, li2024drowzee, chang2024detecting}. Therefore, effectively identifying the semantic differences of entities is an effective solution for detecting hallucinations. Based on this, we select the Wikipedia dataset, denoted as \( W \), which contains several independent article paragraphs \(\{w_1, w_2, ..., w_n\}\). From each article \( w_i \), we extract key entities (such as names, years, numbers, and other important terms), and truncate each article \( w_i \) based on the positions of the extracted entities \( e_i \) (excluding the beginning of each sentence). The truncated document is then fed into the LLM with the prompt: "Below is the opening sentence from a Wikipedia article titled [Title]. Please continue the passage from where the sentence ends. [First sentence of that article]." We choose LLaMA2-7b-CHAT to generate a new piece of text based on the prompt, which we define as \( t_i \), as shown in lines 1-7 of Algorithm \ref{al:2}.

To detect whether the newly generated text contains hallucinations, we perform entity extraction on the same truncated portion of \( t_i \) and the original text \( w_i \), resulting in \( n_i \). We then compare the extracted entities \( n_i \) with the original entities \( e_i \) to determine if a hallucination has occurred. However, entities often exist in multiple forms, such as abbreviations, variations in naming conventions, or different word orders. Therefore, simple direct matching may not capture all entity-level errors. To address this, we combine manual evaluation with the cosine similarity comparison between the vector representations of the extracted entities and the originally selected entities. In this way, we can assess whether two entities are semantically equivalent. If the entities are semantically similar (e.g., "USA" and "United States of America" are considered the same), a hallucination is not detected. Otherwise, the LLM-generated text is labeled as a hallucination. This is specifically described in lines 8-18 of step 2 in Algorithm \ref{al:2}.

Based on the above steps, we construct a real-time hallucination detection dataset \( D_i \), which includes \((w_i, e_i, t_i, n_i, H_i)\), where \( H_i \) represents the hallucination label, with 0 indicating a hallucination and 1 indicating no hallucination. For \( D_i \), we train a multi-layer perceptron (MLP) model.

\begin{equation}
P = \text{MLP}(\mathbf{D}_i\left\{\mathbf{w}_i, \mathbf{e}_i, \mathbf{t}_i, \mathbf{n}_i, \mathbf{H}_i\right\} + b),
\end{equation}
where \(b\) is the bias vector.

\subsection{How many training samples are generated from Wikipedia?}

In this subsection, we mainly explain how much data is constructed to train these two classifiers: Early Detection and Real-time Detection.

\subsubsection{The number of training samples of Early Detection}

We used LLaMA2-7B to construct 5201 data from Wikipedia, with a training set and a test set ratio of 0.8:0.2, to train and validate Early Detection. We use an NVIDIA A100 GPU, which processes for about half an hour.

\subsubsection{The number of training samples of Real-time Detection}

We constructed the training set for Real-time Detection from Wikipedia by means of the LLaMA2-7B model. Our dataset comprises 6,000 training samples, 1,000 validation samples, and 1,304 test samples. We use an NVIDIA A100 GPU, which processes for about one hour.

\section{Experimental Setups}
\label{sec:appendix2}
\subsection{Datasets}

We evaluate our approach using four multi-hop benchmark datasets and three single-hop benchmark datasets, each designed to assess different aspects of model reasoning and performance:

\textbf{\textit{a)}} 2WikiMultihopQA \cite{ho2020constructing}: This dataset is specifically designed to test a model's ability to perform chain-of-thought (CoT) reasoning. It challenges the model to generate answers by integrating information from multiple Wikipedia articles, requiring the model to navigate through several hops of information to synthesize a final response. It has 1000 examples.

\textbf{\textit{b)}} HotpotQA \cite{yang2018hotpotqa}: Similar to 2WikiMultihopQA, this dataset focuses on questions that necessitate information retrieval from multiple documents. It evaluates the model's capacity to engage in CoT reasoning, requiring it to identify and connect different pieces of evidence scattered across diverse passages to formulate a coherent and accurate answer. It has 1000 examples.

\textbf{\textit{c)}} IIRC \cite{abdelsalam2021iirc}: The IIRC dataset is aimed at evaluating reading comprehension and advanced synthesis. Questions in this dataset require the model to integrate information from multiple documents, often spanning several passages, and demand a deep level of synthesis to derive accurate answers from complex, interconnected pieces of text. It has 954 examples.

\textbf{\textit{d)}} StrategyQA \cite{geva2021did}: This dataset is designed to assess commonsense reasoning by posing questions that involve implicit strategies. To answer these questions, models must generate CoT reasoning steps and leverage abstract thinking, making it a strong test for evaluating a model's commonsense knowledge and its ability to navigate reasoning processes that are not explicitly stated. It has 1000 examples.

\textbf{\textit{e)}} NaturalQuestions \cite{kwiatkowski2019natural}: 
NaturalQuestions is a large-scale open-domain question answering dataset constructed from real user queries issued to Google Search. Each question is paired with a Wikipedia page, and annotated with both long answers (typically paragraphs) and short answers (typically named entities or phrases). The dataset is challenging due to the natural, free-form nature of the questions and the need for deep contextual understanding, making it a strong benchmark for evaluating information retrieval and reading comprehension models.

\textbf{\textit{f)}} TriviaQA \cite{joshi2017triviaqa}: TriviaQA is a question answering dataset collected from trivia websites, featuring naturally occurring questions written by trivia enthusiasts. Each question is associated with multiple evidence documents retrieved from Wikipedia and the web, with answers annotated within these documents. The dataset covers a broad range of topics and requires models to perform reasoning across long and sometimes noisy contexts, providing a robust test for reading comprehension in real-world settings.

\textbf{\textit{g)}} SQuAD \cite{rajpurkar2018know}: The Stanford Question Answering Dataset (SQuAD) is one of the most widely used benchmarks in QA research. In version 2.0, it combines answerable and unanswerable questions, requiring models to either extract a precise span from a given Wikipedia paragraph or determine that no answer is present. SQuAD emphasizes fine-grained language understanding, span-level prediction, and the ability to distinguish between answerable and unanswerable contexts.

\subsection{Evaluation Metrics}
We employed a comprehensive set of metrics to evaluate the performance of DioR and other RAG methods on different datasets and retrieval methods. These metrics can be broadly categorized into two groups: \textit{effectiveness metrics} and \textit{efficiency metrics}. Specifically as follows:

Effectiveness metrics aimed to measure the accuracy and quality of the generated answers. We evaluated the models using 
Exact Match (EM), F1 Score, Precision (Pre.), and Recall (Rec.). The EM metric measures the proportion of model-generated answers that exactly match the reference answers, providing a strict assessment of answer accuracy. The F1 Score calculates the harmonic mean of 
precision and recall, offering a balanced evaluation of answer quality. We assessed the precision and recall of the models to evaluate their ability to produce accurate responses.

Efficiency metrics focused on assessing the computational resources and time complexity associated with each model. We measured the Retrieve Count (Rc) for the number of documents retrieved during the retrieval process, which measures the ability of the model to efficiently gather relevant information. Furthermore, we calculated the Generate Count (Gc) for the number of times generated answers are produced, reflecting the 
model's capacity for generating responses. The Hallucinated Count (Hc) metric assesses the number of hallucinated content 
occurrences, where the model generates responses that are not grounded in the input text or retrieved documents. Finally, we measured the Token Count (Tc) and Sentence Count (Sc) to evaluate the verbosity and response length of the models. Due to the fact that LLM is generated more than once, multiple RAGs may be performed, resulting in cumulative statistics of Tc and Sc indicators.

\subsection{Implementation}

We compare our method with seven RAG baselines, including SEAKR \cite{yao2024seakr}, RaDIO \cite{Zhu_Guo_Shi_Chen_DeMeo_2025}, DRAGIN \cite{DBLP:conf/acl/SuTA0024}, wo-RAG \cite{DBLP:conf/acl/SuTA0024}, SR-RAG \cite{DBLP:conf/acl/SuTA0024}, FL-RAG \cite{borgeaud2022improving, ram2023context}, FS-RAG \cite{Trivedi2022}, and FLARE \cite{jiang2023active, DBLP:conf/acl/SuTA0024}. wo-RAG represents a setting where no retrieval-augmented generation (RAG) is applied. Other methods are already introduced in the related work section. Notably, wo-RAG and SR-RAG are single-round RAG methods, while the remaining techniques employ a multi-round retrieval strategy.

For our experiments, DRAGIN serves as the baseline (we express it as Base), referenced throughout this work. We employ two fine-tuned large language models as the underlying backbone: LLaMA2-7B CHAT \cite{touvron2023llama} and Qwen2.5-7B \cite{yang2024qwen2}, both optimized for dialogue-oriented tasks.

Three retrieval strategies are employed: BM25 \cite{lv2011adaptive}, SBERT \cite{wang2020sbert}, and SGPT \cite{muennighoff2022sgpt}. Specifically as follows, 

1) BM25: It is a classic information retrieval algorithm based on the idea of TF-IDF (Word Frequency Inverse Document Frequency), but has been improved to consider factors such as document length. It can quantitatively evaluate the relevance between documents and queries, help determine the most relevant documents, and thus improve retrieval performance.

2) SGPT: It is a semantic search sentence embedding method based on GPT model. It utilizes the powerful semantic understanding ability of the GPT model and proposes two architectures: dual encoder (SGPT-BE) and cross encoder (SGPT-CE). SGPT-BE generates sentence representations by fine-tuning the bias tensor of the GPT model and using position weighted average pooling.

3) SBERT: It is a variant based on BERT, designed specifically for generating sentence embeddings, mainly used for tasks such as calculating semantic similarity, text clustering, and information retrieval. It optimizes the generation of semantic embeddings, making it more efficient to calculate the similarity between sentence pairs. SBERT solves the problem of low efficiency in semantic similarity calculation of the original BERT model. By introducing sentence embeddings, each sentence only needs to be encoded once, greatly improving efficiency.

The experiments are conducted on NVIDIA A100 80GB. For hyperparameter settings, we set the maximum generated sequence length to 256 tokens, with a retrieval top-$k$ value of 3 and max retrieval times of 5, and selected the top 25 passages to ensure high-quality responses. We used datasets containing 1000 data points each to ensure robust and reliable results. We processed approximately 12,000 samples on four A100 GPUs. The total runtime was around 18 hours, with an average overall response time of 5.4 seconds per sample.

\section{More Experimental Results}
\label{sec:appendix3}

\begin{table*}[h]
\centering
\caption{Compare the efficiency of various retrieval and generation methods in base and DioR across four different datasets.}
\resizebox{\textwidth}{!}{%
\begin{tabular}{l|ccccccccccccc}
\toprule
\multirow{2}{*}{\textbf{Matrix}} & \multicolumn{2}{c}{\textbf{2WikiMultihopQA}} & \multicolumn{2}{c}{\textbf{HotpotQA}} & \multicolumn{2}{c}{\textbf{IIRC}} & \multicolumn{2}{c}{\textbf{StrategyQA}} \\
\cmidrule(r){2-3} \cmidrule(r){4-5} \cmidrule(r){6-7} \cmidrule(r){8-9}
 & Base & \textbf{DioR} & Base & \textbf{DioR} & Base & \textbf{DioR} & Base & \textbf{DioR} \\
\midrule
Rc (BM25) & 1.018 & \textbf{3.006} & 2.832 & \textbf{3.033} & 3.696 & \textbf{3.019} & 4.422 & \textbf{3.173} \\
Rc (SGPT) & 3.602 & \textbf{1.932} & 3.002 & \textbf{2.053} & 3.002 & \textbf{2.023} & 4.305 & \textbf{2.213} \\
Rc (SBERT) & 1.804 & \textbf{4.006} & 0.974 & \textbf{4.060} & 1.534 & \textbf{4.018} & 2.390 & \textbf{4.258} \\
\midrule
Gc (BM25) & 2.038 & \textbf{3.006} & 6.372 & \textbf{3.037} & 7.431 & \textbf{3.022} & 8.849 & \textbf{3.206} \\
Gc (SGPT) & 7.208 & \textbf{2.021} & 6.007 & \textbf{2.092} & 6.009 & \textbf{2.074} & 8.613 & \textbf{2.280} \\
Gc (SBERT) & 3.982 & \textbf{2.033} & 2.725 & \textbf{2.098} & 3.206 & \textbf{2.062} & 5.601 & \textbf{2.206} \\
\midrule
Hc (BM25) & 1.018 & \textbf{1.003} & 2.832 & \textbf{2.017} & 3.696 & \textbf{2.010} & 4.422 & \textbf{2.093} \\
Hc (SGPT) & 3.602 & \textbf{1.005} & 3.002 & \textbf{1.046} & 3.002 & \textbf{1.035} & 4.305 & \textbf{1.134} \\
Hc (SBERT) & 1.804 & \textbf{1.014} & 0.974 & \textbf{1.046} & 1.534 & \textbf{1.030} & 2.390 & \textbf{1.098} \\
\midrule
Tc (BM25) & 523.763 & \textbf{772.585} & 694.887 & \textbf{391.962} & 959.080 & \textbf{776.529} & 894.252 & \textbf{323.321} \\
Tc (SGPT) & 1852.543 & \textbf{519.481} & 775.129 & \textbf{270.201} & 775.328 & \textbf{533.008} & 870.086 & \textbf{229.660} \\
Tc (SBERT) & 537.053 & \textbf{522.566} & 216.537 & \textbf{270.896} & 497.527 & \textbf{529.970} & 325.277 & \textbf{222.434} \\
\midrule
Sc (BM25) & 36.530 & \textbf{47.589} & 34.372 & \textbf{21.491} & 47.099 & \textbf{42.073} & 59.893 & \textbf{23.066} \\
Sc (SGPT) & 123.930 & \textbf{32.572} & 40.494 & \textbf{14.887} & 36.826 & \textbf{29.014} & 60.191 & \textbf{16.984} \\
Sc (SBERT) & 30.908 & \textbf{32.544} & 10.629 & \textbf{14.575} & 25.094 & \textbf{29.376} & 23.592 & \textbf{16.293} \\
\bottomrule
\end{tabular}
}
\label{tab:eff}
\end{table*}

\subsection{Efficiency Comparison Experiment}

As shown in Table \ref{tab:eff}, We discuss the effectiveness of DioR from the perspective of efficiency indicators.

DioR demonstrates a more efficient retrieval and generation process. For example, in 2WikiMultihopQA, DioR retrieves significantly more documents, reaching 3.006, while reducing hallucinated generations to just 1.003. The same findings can also be observed from SBERT.

In HotpotQA, DioR retrieves more documents with BM25 (3.033 vs. 2.832) but generates fewer responses (Gc 3.037 vs. Base 6.372), leading to more computationally efficient performance.

Additionally, DioR reduces hallucinated content across all retrieval methods. In 2WikiMultihopQA (BM25), DioR's Hc is 1.003, much lower than Base’s 1.018, reflecting its improved ability to ground responses and minimize irrelevant information. This trend is consistent across datasets, with DioR generally exhibiting fewer hallucinations than the Base method.

In terms of verbosity, DioR produces slightly longer responses in some cases, such as 2WikiMultihopQA (Tc 772.585 vs. Base 523.763), but this is balanced by its ability to generate more precise answers with fewer sentences (Sc 47.589 vs. 36.530 with BM25). In datasets like StrategyQA, DioR achieves a more concise output (Tc 323.321 vs. 894.252 for Base) without sacrificing accuracy, showing its flexibility in adjusting verbosity based on the task requirements.

DioR outperforms the Base method by retrieving more relevant documents, generating fewer responses, and reducing hallucinations, all while maintaining a balanced verbosity. These improvements result in a more efficient and effective approach to complex question-answering tasks.

\subsection{Compare With Single-hop Datasets}

\begin{table}[ht]
\caption{Performance of various models on NaturalQuestions, TriviaQA, and SQuAD using LLaMA2-7B and BM25. Metrics shown are Exact Match (EM) and F1 scores.}
\centering
\resizebox{\linewidth}{!}{%
\begin{tabular}{lcccccc}
\toprule
\textbf{Model} & \multicolumn{2}{c}{\textbf{NQ}} & \multicolumn{2}{c}{\textbf{TriviaQA}} & \multicolumn{2}{c}{\textbf{SQuAD}} \\
\cmidrule(lr){2-3} \cmidrule(lr){4-5} \cmidrule(lr){6-7}
 & \textbf{EM} & \textbf{F1} & \textbf{EM} & \textbf{F1} & \textbf{EM} & \textbf{F1} \\
\midrule
DioR      & 26.2 & 35.9 & 52.3 & 61.9 & 21.5 & 27.8 \\
CoT       & 13.4 & 18.7 & 42.6 & 48.6 & 8.7  & 13.6 \\
Self-RAG  & 32.3 & 40.2 & 21.2 & 37.9 & 5.1  & 18.3 \\
FLARE     & 25.3 & 35.9 & 51.5 & 60.3 & 19.4 & 28.3 \\
DRAGIN    & 23.2 & 33.2 & 42.0 & 62.3 & 18.7 & 28.7 \\
RaDIO     & 24.6 & 34.0 & 48.3 & 61.8 & 20.4 & 27.5 \\
\bottomrule
\end{tabular}}
\label{tab:single}
\end{table}

As shown in Table \ref{tab:single}, we conducted experiments on three single-hop datasets: NaturalQuestions, TriviaQA, and SQuAD. From the results, we observed that DioR achieves certain advantages in terms of EM and F1 scores in most cases. Compared to other dynamic RAG methods such as FLARE, DRIGIN, and RaDIO, our approach consistently demonstrates a certain level of improvement.

\subsection{Compare with other LLM backbone}

\begin{table*}[htbp]
\centering
\label{tab:multihop_metrics}
\caption{Performance Metrics for Multiple Datasets using Qwen2.5-7B}
\resizebox{\textwidth}{!}{
\begin{tabular}{lcccccccccccc}
\hline
\textbf{Metric} & \multicolumn{3}{c}{\textbf{2WikiMultihopQA}} & \multicolumn{3}{c}{\textbf{HotpotQA}} & \multicolumn{3}{c}{\textbf{IIRC}} & \multicolumn{3}{c}{\textbf{StrategyQA}} \\
\textbf{} & Base & RaDIO & DioR & Base & RaDIO & DioR & Base & RaDIO & DioR & Base & RaDIO & DioR \\
\hline
EM (BM25)       & 0.165  & 0.178  & 0.214  & 0.111  & 0.124  & 0.163  & 0.1488 & 0.1594 & 0.1719 & 0.768  & 0.776  & 0.789 \\
F1 (BM25)       & 0.246  & 0.2641 & 0.2902 & 0.1951 & 0.2236 & 0.2548 & 0.1851 & 0.2039 & 0.2086 & 0.768  & 0.776  & 0.789 \\
Precision (BM25) & 0.259  & 0.2704 & 0.2913 & 0.1931 & 0.2261 & 0.2605 & 0.1899 & 2.0645 & 0.2145 & 0.768  & 0.776  & 0.789 \\
Recall (BM25)   & 0.245  & 0.2556 & 0.3032 & 0.3003 & 0.3023 & 0.3052 & 0.1939 & 0.2036 & 0.2153 & 0.768  & 0.776  & 0.789 \\
\hline
\end{tabular}
}

\resizebox{\textwidth}{!}{
\begin{tabular}{lcccccccccccc}
\hline
\textbf{Metric} & \multicolumn{3}{c}{\textbf{2WikiMultihopQA}} & \multicolumn{3}{c}{\textbf{HotpotQA}} & \multicolumn{3}{c}{\textbf{IIRC}} & \multicolumn{3}{c}{\textbf{StrategyQA}} \\
\textbf{} & Base & RaDIO & DioR & Base & RaDIO & DioR & Base & RaDIO & DioR & Base & RaDIO & DioR \\
\hline
Rc (BM25)       & 1.866  & 1.015  & 2.036  & 2.904  & 3.024  & 2.524  & 1.9004 & 2.1962 & 2.0922 & 1.776  & 3.026  & 2.638 \\
Gc (BM25)       & 3.737  & 2.301  & 2.04   & 4.386  & 3.985  & 2.553  & 3.8050 & 3.136  & 2.0953 & 4.06   & 2.897  & 2.703 \\
Hc (BM25)       & 1.866  & 1.135  & 1.019  & 2.094  & 1.765  & 1.263  & 1.9004 & 1.693  & 1.0471 & 1.776  & 1.471  & 1.321 \\
Tc (BM25)       & 856.514 & 521.979 & 442.793 & 302.894 & 200.398 & 194.395 & 868.561 & 697.563 & 515.927 & 239.286 & 210.397 & 171.608 \\
Sc (BM25)       & 54.754 & 36.474 & 28.751 & 19.146 & 13.562 & 12.273 & 48.010 & 36.854 & 30.769 & 17.845 & 14.672 & 12.487 \\
\hline
\end{tabular}
}
\label{tab:Qwen}
\end{table*}

We have supplemented our results with experiments using Qwen2.5-7B, comparing performance from both effectiveness and efficiency perspectives. Our results on four knowledge-intensive datasets further validate the effectiveness of DioR. Specifically see Table \ref{tab:Qwen}.

\subsection{More effciency ablation study}

\begin{table*}[htbp]
\centering
\caption{Comparison of Gc and Hc Metrics Across Datasets and Retrieval Methods (w/o ED vs DioR)}
\resizebox{0.9\textwidth}{!}{
\begin{tabular}{lcccccccc}
\hline
\textbf{Metric} & \multicolumn{2}{c}{\textbf{2WikiMultihopQA}} & \multicolumn{2}{c}{\textbf{HotpotQA}} & \multicolumn{2}{c}{\textbf{IIRC}} & \multicolumn{2}{c}{\textbf{StrategyQA}} \\
\textbf{} & w/o ED & DioR & w/o ED & DioR & w/o ED & DioR & w/o ED & DioR \\
\hline
Gc (BM25)   & 3.165 & 3.006 & 3.574 & 3.037 & 3.566 & 3.022 & 3.406 & 3.206 \\
Gc (SBERT)  & 3.356 & 2.021 & 2.357 & 2.092 & 2.431 & 2.074 & 2.862 & 2.280 \\
Gc (SGPT)   & 2.833 & 2.033 & 2.114 & 2.098 & 3.012 & 2.062 & 2.232 & 2.206 \\
Hc (BM25)   & 1.015 & 1.003 & 2.041 & 2.017 & 2.209 & 2.010 & 2.131 & 2.093 \\
Hc (SBERT)  & 1.035 & 1.014 & 1.075 & 1.046 & 1.084 & 1.030 & 1.268 & 1.098 \\
Hc (SGPT)   & 1.016 & 1.005 & 1.304 & 1.046 & 1.242 & 1.035 & 1.165 & 1.134 \\
\hline
\end{tabular}
}
\label{tab:gc_hc_comparison}
\end{table*}

Our efficiency analysis further demonstrates that, compared to the baseline (DRAGIN), DioR effectively minimizes hallucinations, improves the conciseness of generated content, and optimizes inference time. Consequently, DioR not only achieves accuracy improvements over existing state-of-the-art RAG methods while maintaining answer quality but also significantly reduces computational overhead. The table \ref{tab:gc_hc_comparison} below presents an efficiency experiment that validates the effectiveness of our approach in improving computational efficiency. (Early-Detection: ED)

\subsection{Case Study}

\begin{table*}[t]
\centering
\caption{Comparison between the Base method and DioR methods in terms of Hc, Sc, and Gc for a case study. The table highlights specific examples, their predictions and outcomes, and the metrics associated with each method. A checkmark ( $\checkmark$ ) denotes correct outcomes, while a cross ( $\times$ ) indicates incorrect outcomes.}
\setlength{\tabcolsep}{2pt}
\resizebox{\textwidth}{!}{%
\begin{tabular}{c|m{8cm}|c|c|c|c|m{8cm}|c|c|c|c}
\hline
\textbf{Example} & \multicolumn{2}{c|}{\textbf{Base}}& \multicolumn{3}{c|}{\textbf{Num.}} & \multicolumn{2}{c|}{\textbf{DioR}} & \multicolumn{3}{c}{\textbf{Num.}}\\
\hline
 & \textbf{Prediction} & \textbf{Outcome} & \textbf{Hc} & \textbf{Sc} & \textbf{Gc} & \textbf{Prediction} & \textbf{Outcome} & \textbf{Hc} & \textbf{Sc} & \textbf{Gc} \\
\hline
1 & The text states that Howard worked at the train station, which is a place of business. Businesses are typically open during the summer. Thus, Santa Claus works during summer. So the answer is yes. & $\times$ & 2 & 16 & 5 & Santa Claus works during winter. Winter is from December to February. Summer is from June to August. Thus, Santa Claus does not work during summer. So the answer is no. & $\checkmark$ & 1 & 15 & 2 \\
\hline
2 & An uninsured person would be more likely than an insured person to decline a CT scan. Reasoning:\newline 1. The diagnostic accuracy of coronary CT angiography (CCTA) is similar to that of invasive coronary angiography via cardiac catheterization (ICA). (1)\newline 2. Both CCTA and ICA yield similar prognostic information, such as stenosis severity and extent of coronary artery disease. & $\times$ & 4 & 46 & 8 & The cost of a CT scan is \$500. If a person has insurance, the person's insurance company pays for the CT scan. If a person does not have insurance, the person pays for the CT scan. Thus, if a person does not have insurance, the person is more likely than an insured person to decline a CT scan. So the answer is yes. & $\checkmark$ & 1 & 16 & 2 \\
\hline
3 & 1. The text states that the Christmas tree was a 40-foot Colorado Blue Spruce donated by the Loving Shepherd Lutheran Church and School.\newline 2. Deciduous trees are trees that lose their leaves in the fall.\newline 3. Colorado Blue Spruce trees are evergreen, which means they keep their needles year-round. Thus, the Christmas tree is not dissimilar to deciduous trees. So the answer is no. & $\times$ & 6 & 59 & 13 & Christmas trees are evergreen. Evergreen trees are not deciduous. Thus, Christmas trees are dissimilar to deciduous trees. So the answer is yes. & $\checkmark$ & 1 & 11 & 2 \\
\hline
4 & No.\newline Reasoning:\newline The passage states that the official language of Guernsey is English, but it also mentions that French was the official language until 1948. This implies that the language used in Guernsey is not rooted in English. Therefore, the answer is no. & $\times$ & 2 & 15 & 5 & Saint Vincent and the Grenadines is an independent country. The official language of an independent country is the language of the country. Thus, the language used in Saint Vincent and the Grenadines is rooted in English. So the answer is yes. & $\checkmark$ & 1 & 10 & 2 \\
\hline
5 & Genghis Khan lived in the 13th century. Julius Caesar lived in the 1st century. Today is in the 21st century. The distance between the 13th century and the 21st century is 890 years. There are 890 years between the number of people alive today and Genghis Khan. Thus, more people today are related to Genghis Khan than Julius Caesar. So the answer is yes. & $\checkmark$ & 5 & 33 & 11 & The world population is about 7 billion. Genghis Khan lived about 800 years ago. Julius Caesar lived about 2000 years ago. So, more people today are related to Genghis Khan than Julius Caesar. Thus, the answer is yes. & $\checkmark$ & 1 & 12 & 2 \\
\hline
6 & No. Explanation: 1. Hades is a character from Greek mythology. 2. The Disney Channel is a television network. 3. Musical movies are movies with music. 4. Hades does not appear in any Disney Channel musical movies. Therefore, the answer is no. & $\times$ & 3 & 27 & 6 & Hades appears in the Disney Channel musical movie \"Hercules\". Thus, the answer is yes.  & $\checkmark$ & 2 & 18 & 3 \\
\hline
7 & Someone theoretically could use an armadillo as a shield. Reasoning: 1. Armadillos have a hard carapace that protects their body. 2. The carapace is flexible, allowing the armadillo to roll into a ball for defense. 3. The armadillo's small size makes it easier to hold and use as a shield. 4. The armadillo's curved claws could be used to grip onto an attacker's shield. & $\checkmark$ & 7 & 70 & 14 & An armadillo is a mammal. Mammals can use their bodies as shields. Thus, someone theoretically could use an armadillo as a shield. So the answer is yes.  & $\checkmark$ & 2 & 25 & 3 \\
\hline
8 & Mediterranean Sea creatures are not known to kill humans. Steve Irwin was killed by a stingray. Thus, no Mediterranean Sea creature killed Steve Irwin. So the answer is no. & $\times$ & 2 & 11 & 3 & Steve Irwin died when a stingray barb pierced his chest. The stingray lives in the Mediterranean Sea. Thus, the Mediterranean Sea creature that killed Steve Irwin was a stingray. So the answer is yes.  & $\checkmark$ & 1 & 13 & 3 \\
\hline
9 & The NYPD police officer age requirement is 21 years old. A Generation Y member is someone born between 1980 and 1994. The youngest Generation Y member is 26 years old, which is older than the NYPD police officer age requirement. Thus, a Generation Y member does not satisfy the NYPD police officer age requirement. So the answer is no. & $\times$ & 3 & 24 & 7 & The NYPD age requirement is 21. Generation Y is 1980 and 1994. 1994 is the year Generation Y ends. Thus, a Generation Y member could satisfy the NYPD age requirement. So the answer is yes.   & $\checkmark$ & 2 & 14 & 3 \\
\hline
10 & Someone with back pain could enjoy picking strawberries. The reason is that picking strawberries is a sedentary activity. Sedentary activities do not strain the back. Thus, picking strawberries would not cause back pain. & $\times$ & 3 & 23 & 6 & Strawberry picking typically involves bending, twisting, and lifting weights, which may cause discomfort for patients with back pain.   & $\checkmark$ & 2 & 12 & 3 \\
\hline
\end{tabular}
}
\label{table:comparison}
\end{table*}

As shown in Table \ref{table:comparison}, we present a case study comparing the Base method and DioR on five example questions, evaluating their ability to generate accurate answers while minimizing hallucinations (Hc), reducing verbosity (Sc), and optimizing response generation (Gc). For each example, we assess the correctness of the predictions and the associated metrics.

\begin{enumerate}
    \item \textbf{Does Santa Claus work during summer?} \\
    The Base method incorrectly asserts that Santa Claus works during summer based on flawed reasoning, leading to a hallucination (Hc = 2) and generating 16 sentences (Sc = 16). In contrast, DioR correctly deduces that Santa Claus works during winter, producing a correct answer with only one hallucination (Hc = 1) and 15 sentences (Sc = 15), but with a lower number of generations (Gc = 2), indicating a more efficient response.

    \item \textbf{Would an uninsured person be more likely than an insured person to decline a CT scan?} \\
    The Base method fails to reason effectively, resulting in incorrect reasoning and an answer that was not grounded in the text, with 4 hallucinations (Hc = 4) and 46 sentences (Sc = 46). DioR, however, produces a grounded and correct response with minimal hallucination (Hc = 1) and more concise reasoning (Sc = 16) while generating only 2 responses (Gc = 2).

    \item \textbf{Are Christmas trees dissimilar to deciduous trees?} \\
    The Base method incorrectly concludes that Christmas trees are similar to deciduous trees, producing a large number of hallucinations (Hc = 6) and an extensive response (Sc = 59). DioR, however, correctly answers that Christmas trees are evergreen and not deciduous, with significantly fewer hallucinations (Hc = 1) and a more succinct response (Sc = 11), generating only 2 answers (Gc = 2), indicating better efficiency and correctness.

    \item \textbf{Is the language used in Saint Vincent and the Grenadines rooted in English?} \\
    The Base method produces a false response, relying on incorrect reasoning, resulting in 2 hallucinations (Hc = 2) and 15 sentences (Sc = 15). DioR provides the correct answer, grounded in more accurate reasoning, with only 1 hallucination (Hc = 1) and a more concise response (Sc = 10), generating 2 answers (Gc = 2).

    \item \textbf{Are more people today related to Genghis Khan than Julius Caesar?} \\
    Both methods provide a correct answer, but the Base method's reasoning leads to a higher number of hallucinations (Hc = 5) and a larger response (Sc = 33). DioR correctly answers with more concise reasoning, fewer hallucinations (Hc = 1), and shorter response length (Sc = 12), generating only 2 answers (Gc = 2), demonstrating higher efficiency.

    \item \textbf{Does Hades appear in a Disney Channel musical movie?} \\

    The Base method incorrectly concludes that Hades does not appear in a Disney Channel musical movie, producing a large number of hallucinations  and an extensive response. In contrast, DioR correctly answers, with significantly fewer hallucinations (Hc = 2) and a more succinct response (Sc = 18), indicating better efficiency and correctness

    \item \textbf{Could someone theoretically use an armadillo as a shield?} \\

    Although both the Base and DioR methods produce the correct answer that someone theoretically use an armadillo as a shield. However, DioR demonstrates a greater advantage in metrics.
    
    \item \textbf{Did a Mediterranean Sea creature kill Steve Irwin?} \\

    For this question, the Base method incorrectly believes that Steve Irwin was killed by Stingray, so the answer is incorrect. In contrast, DioR can generate fewer hallucination (Hc = 1) to answer correctly.
        
    \item \textbf{Does a Generation Y member satisfy NYPD police officer age requirement?} \\

    The logic of the Base method is confusing. The age requirement for the New York Police Department is 21 years old, while the age range for millennial members is from 26 to 41 years old (based on members born in 1994). Since the minimum age for members of the millennial generation is 26 years old, which is already higher than 21 years old, they clearly meet the age requirements of the New York Police Department. On the contrary, DioR can correctly derive the answer and all indicators are superior to Base.
    
    \item \textbf{Would someone with back pain enjoy picking strawberries?} \\

    The description of the Base method is incorrect. Strawberry picking is not a sedentary activity, but a physical activity that requires frequent bending, twisting, and weightlifting. This activity may cause discomfort for people suffering from back pain. And DioR can output the correct answer, that strawberry picking does require these actions, which may cause discomfort for people with back pain.
    
\end{enumerate}

DioR consistently outperforms the Base method across all five examples by producing correct answers with fewer hallucinations, more efficient responses (lower Gc), and more concise text (lower Sc). These results demonstrate DioR’s superior ability to ground its answers while minimizing unnecessary verbosity, highlighting its advantage in handling complex reasoning tasks.

\begin{figure}[!t]
    \centering
    \includegraphics[width=\linewidth]{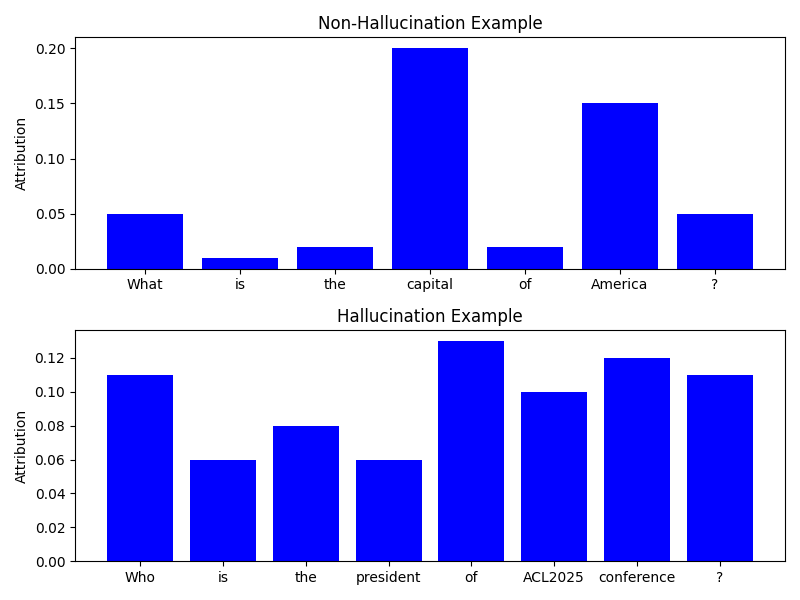}
    \caption{
    An example of an IG attribution score for hallucination and non-hallucination. 
    }
    \label{fig:IG_exp}
\end{figure}

\begin{figure}[!t]
    \centering
    \includegraphics[width=\linewidth]{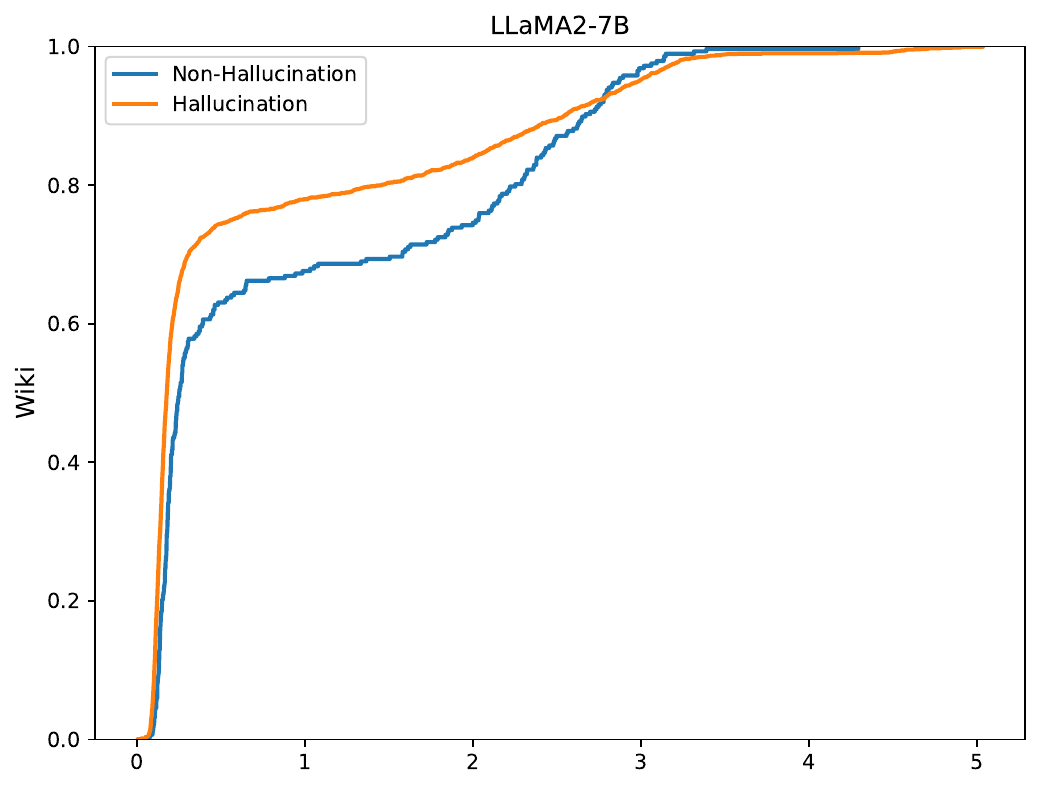}
    \caption{
    The difference in the distribution of the output entropy values when the model generates the correct answer (non-hallucination) and the wrong answer (hallucination).
    }
    \label{fig:ecdf_final}
\end{figure}

\begin{figure}[!t]
    \centering
    \includegraphics[width=\linewidth]{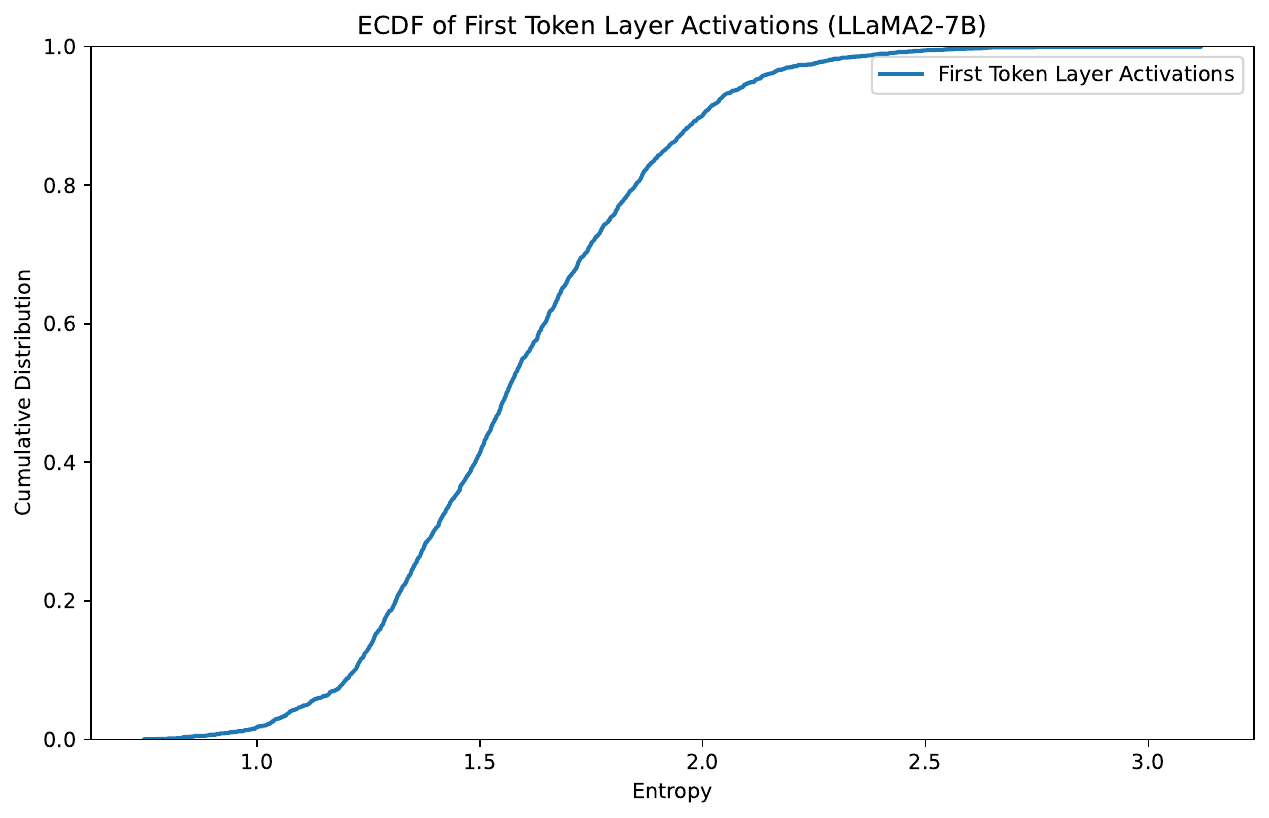}
    \caption{
    The uncertainty distribution of the model when processing the input and generating the first token.
    }
    \label{fig:ecdf_first}
\end{figure}

\begin{figure}[!t]
    \centering
    \includegraphics[width=\linewidth]{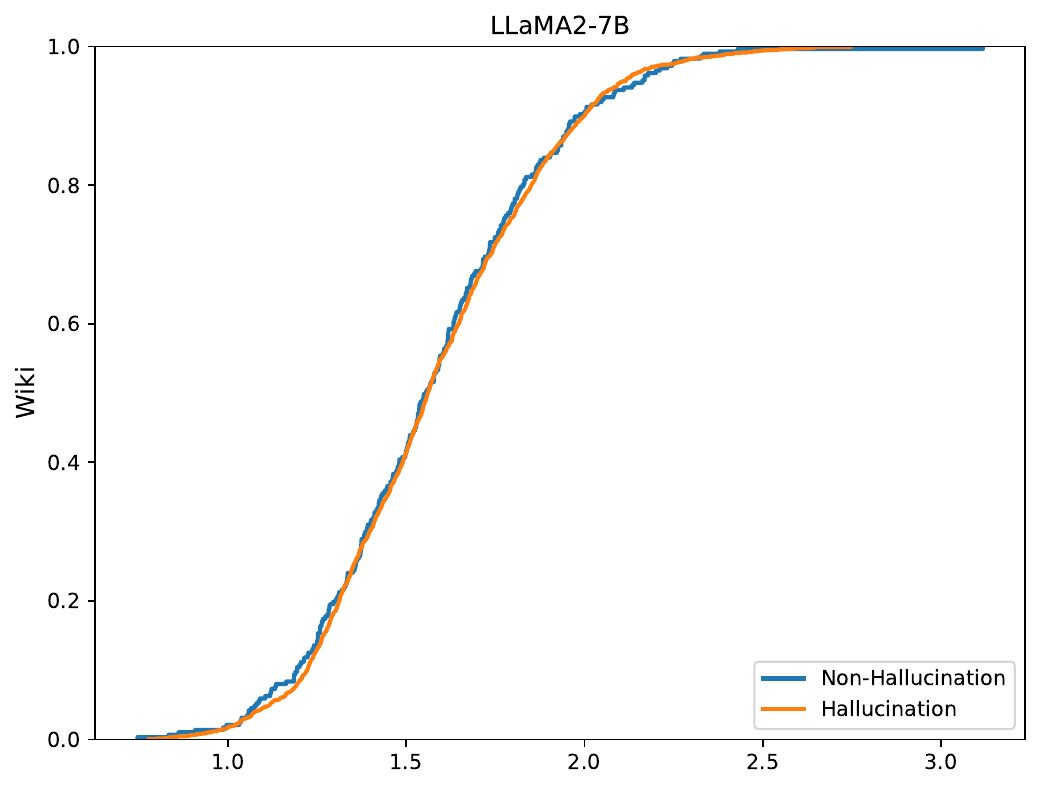}
    \caption{
    The difference in the distribution of input token attribution scores when the model generates correct answers and incorrect answers.
    }
    \label{fig:ecdf_ig}
\end{figure}

\begin{figure}[!t]
    \centering
    \includegraphics[width=\linewidth]{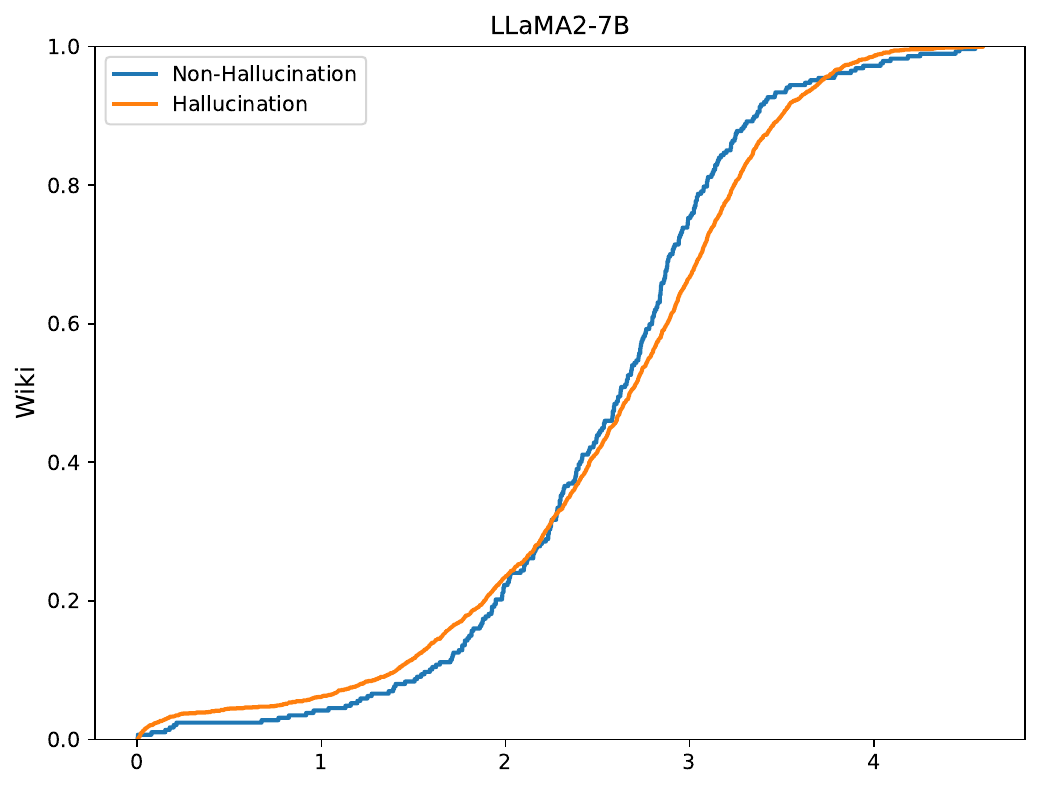}
    \caption{
    The model outputs the difference in softmax probability distribution when generating correct answers and incorrect answers.
    }
    \label{fig:ecdf_softmax}
\end{figure}

\begin{figure}[!t]
    \centering
    \includegraphics[width=\linewidth]{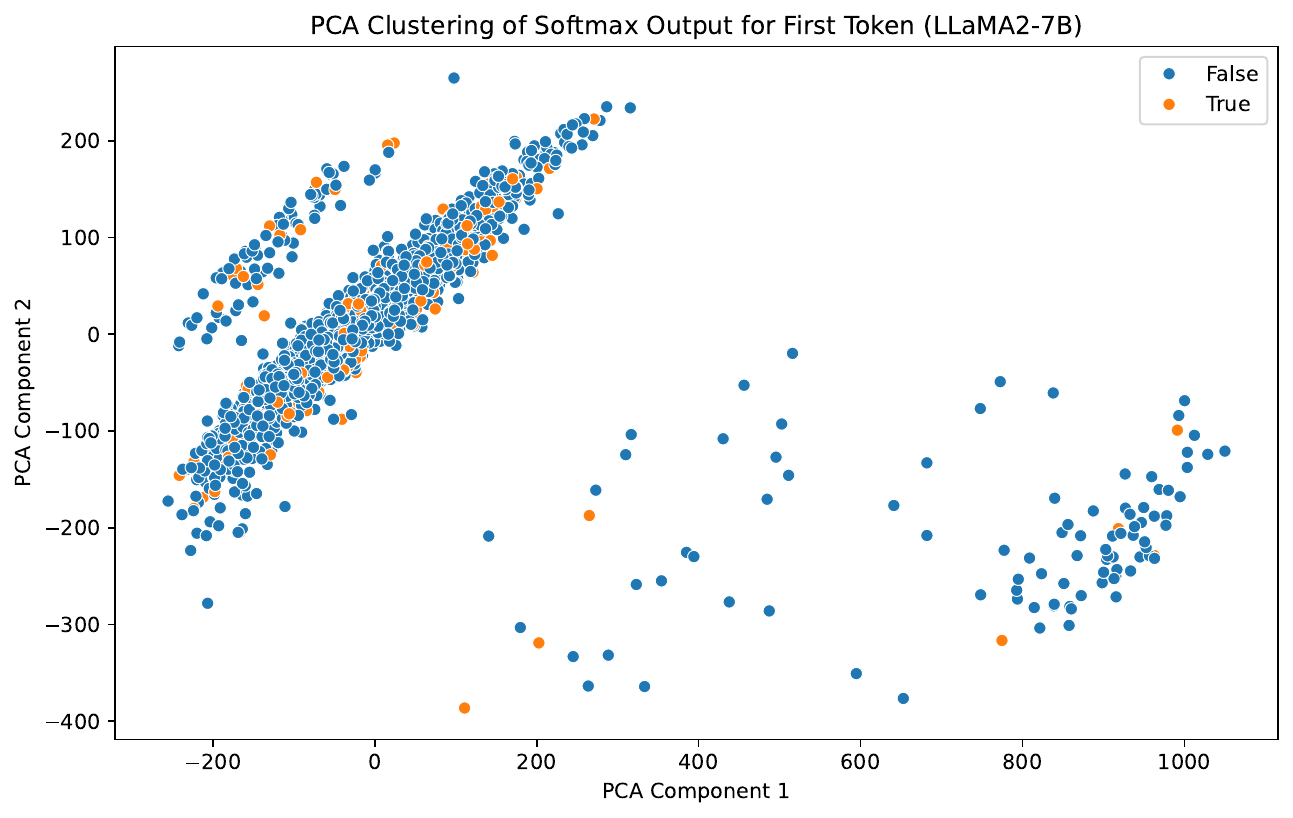}
    \caption{
    The distribution of correct and incorrect answers in the PCA dimensionality reduction space when the model generates the first token.
    }
    \label{fig:softmax_pca}
\end{figure}

\subsection{Effectiveness Analysis of Hallucination Detection Methods}
\label{sec:appendix3.3}

As mentioned earlier, to validate whether our method can effectively determine whether a large language model (LLM) is capable of answering a question, we assess the model’s attention to particular words within the input question. Specifically, we use the Integrated Gradients (IG) method to generate feature attribution, which allows us to quantify the attention distribution of the LLM when processing different questions. This approach enables us to determine whether the model is able to recognize and focus on the key features of a question during the answering process, to assess the level of confidence LLMs have in facing a new problem.

As shown in Figure \ref{fig:IG_exp}. By comparing instances of hallucination and non-hallucination generation, we observe significant differences in the model’s attention distribution between the two cases. In non-hallucination examples, the model accurately identifies and focuses on the key information in the question. In contrast, in hallucination examples, the model’s attention is more dispersed, failing to concentrate on the core words, which leads to incorrect or irrelevant responses. These experimental results indicate that the model's ability to effectively focus on key features directly impacts the accuracy of its answers, thus validating that our method can effectively assess whether an LLM is capable of answering a question before the generation to reduce the probability of hallucination generation (The confidence level of LLMs in answering questions).

We can observe from Figures \ref{fig:ecdf_final}, \ref{fig:ecdf_first}, \ref{fig:ecdf_ig}, \ref{fig:ecdf_softmax}, and \ref{fig:softmax_pca} that the model exhibits noticeable differences when generating hallucinations versus non-hallucinations. The horizontal axis of the first three Figures is the entropy value, the vertical axis is the cumulative distribution function, and the horizontal axis of the fourth Figure is the Softmax probability value. These Figures provide insight into the behavior of the LLaMA2-7B model in generating text, analyzed through various methods such as ECDF, PCA, integrated gradients, and more. 

As shown in Figure \ref{fig:ecdf_final}, the two curves for non-hallucination and hallucination are clearly separated in specific entropy ranges, indicating that the model’s behavior differs significantly between these two scenarios.

Figure \ref{fig:ecdf_first} illustrates the cumulative distribution of entropy values from the first layer of activation values when the model generates the first token. By comparing the distributions of activation values for correct and incorrect answers, we can observe the differences in the model's uncertainty during the generation of the initial word, further revealing how the internal state varies between correct and incorrect answers. A higher cumulative distribution in the higher entropy regions suggests that the model experiences greater uncertainty when generating the first token.

Figure \ref{fig:ecdf_ig} shows the cumulative distribution of entropy values for input token attribution scores generated using the integrated gradients (IG) method. This figure highlights the disparity in the model's attention to input tokens when generating correct and incorrect answers. We can observe a subtle difference in IG scores between scenes that produce hallucinations and those that do not.

Figure \ref{fig:ecdf_softmax} displays the cumulative distribution of the Softmax output probabilities. Higher Softmax probability values reflect greater model confidence in a prediction. We can observe that in the high softmax range, the range of non-hallucinations is greater than that of hallucinations, indicating that the model is more confident in generating non-hallucination scenes. However, in the low softmax range, the model appears less confident and prone to hallucinations.

Lastly, Figure \ref{fig:softmax_pca} presents a two-dimensional scatter plot of the Softmax output after PCA dimensionality reduction for the first token. The distribution of correct and incorrect predictions in the PCA space shows a clear separation between the two, indicating that the Softmax output is significantly different when the model generates correct versus incorrect answers.

These findings validate that our Early-Detection method can effectively assess whether a large language model (LLM) is capable of answering a question before it starts generating text.

We also experimented with the Real-time Detection module to prove its feasibility in hallucination detection. We conduct experiments on LLaMA2-7B to judge the state-of-the-art methods for sentence-level hallucination detection, achieving an AUC of 0.7913, compared to 0.6583 for GPT4-HDM \cite{achiam2023gpt} and 0.7876 for MIND \cite{su2024unsupervised}.

\section{Hallucination Prevention vs. Detection and Regeneration}

\begin{lstlisting}[language=Python, caption={Hallucination Prevention (DioR)}]
function answerWithHallucinationControl(question):
   // Initial detection phase
   confidence = EarlyDetection(question)
   
   // Based on confidence, decide whether to retrieve information upfront
   context = confidence ? "" : retrieveRelevantInformation(question)
   answer = ""
   
   // Token-by-token generation
   while not isEndOfSequence(answer) and len(answer) < max_tokens:
       // Generate next token
       next_token = generateNextToken(question, answer, context)
       
       // Real-time detection for hallucination in the token
       if RealTimeDetection(next_token):
           // Hallucination detected, retrieve relevant documents
           new_context = Retrieve()
           // Continue generation with newly retrieved context
           return continueGeneration(question, answer, new_context)
       
       // Token is valid, add it to the answer
       answer += next_token
   
   return answer
\end{lstlisting}

\begin{lstlisting}[language=Python, caption={Detection and Regeneration}]
function answerWithHallucinationControl(question):
    answer = generateLLMResponse(question)
    isHallucination = detectHallucination(question, answer)
    
    if isHallucination:
        // Discard the hallucinated answer
        retrievedContext = retrieveRelevantInformation(question)
        return generateNewResponse(question, retrievedContext)
    else:
        return answer
\end{lstlisting}

\section{Traditional RAG vs. Dynamic RAG}

Traditional RAG:
Traditional RAG uses a single-round retrieval method to perform a one-time document retrieval operation before generating answers. This method is based on the user's original query, retrieves relevant documents, and simultaneously inputs all retrieved information into a large language model, forming a static contextual window. Due to the fact that the retrieval process is only performed once before the start of generation, the model cannot obtain additional knowledge based on the new information requirements that arise during the generation process, which may result in information gaps or inaccuracies in the answers.

Dynamic RAG:
Dynamic RAG fundamentally changes the way information is obtained and integrated by implementing multiple rounds of retrieval processes. It consists of two key steps:
1) Determining the appropriate retrieval timing during the generation process of the LLM;
2) Formulating queries upon retrieval activation.

\begin{lstlisting}[language=Python, caption={Traditional RAG Pseudocode}]
def traditional_rag(user_query, knowledge_base):
    # 1. Single-round retrieval based on user query
    relevant_documents = retrieve_documents(user_query, knowledge_base)
   
    # 2. Input retrieved documents and user query to LLM
    context = prepare_context(relevant_documents)
   
    # 3. Generate response
    response = generate_response(user_query, context)
   
    return response
\end{lstlisting}

\begin{lstlisting}[language=Python, caption={Dynamic RAG Pseudocode}]
def dynamic_rag(user_query, knowledge_base):
    # Initialize response state
    response_so_far = ""
    llm_state = initialize_llm()
   
    # Iteratively generate response
    while not is_generation_complete(llm_state):
        # 1. Determine if retrieval is needed at current generation step
        if should_retrieve_now(llm_state, response_so_far, user_query):
            # 2. Dynamically construct retrieval query
            retrieval_query = formulate_query(llm_state, response_so_far, user_query)
           
            # 3. Retrieve documents
            relevant_documents = retrieve_documents(retrieval_query, knowledge_base)
           
            # 4. Update LLM state with new context
            llm_state = update_llm_with_context(llm_state, relevant_documents)
       
        # 5. Continue generating the next part of the response
        next_token, llm_state = generate_next_token(llm_state)
        response_so_far += next_token
   
    return response_so_far
\end{lstlisting}

\section{The prompt words input to LLM after removing the ``Post-retrieval'' component in the ablation experiment}
\label{sec:appendix3.4}

After removing the ``Post-retrieval'' component in the ablation experiment, the RAG prompt template is as follows:

{
\noindent \textbf{External Knowledge:}
\textbf{[1] $K_{i1}$} \quad
\textbf{[2] $K_{i2}$} \quad
\textbf{...}

\noindent \textbf{Using the external knowledge provided above, please answer the following question:} \quad

\noindent \textbf{Question:} \textbf{[Ques.] }  \quad

\noindent \textbf{Answer:} \textbf{Insert truncated output [] and additional relevant details here}

}

\end{document}